\documentclass[10pt,journal,compsoc]{IEEEtran}

\usepackage{tikz}
\usepackage{comment}
\usepackage{amsmath,amssymb}
\usepackage{bm}
\usepackage{color}
\usepackage{xcolor}
\usepackage{graphicx}
\usepackage{mathtools}
\usepackage{siunitx}
\usepackage{subcaption}
\usepackage{booktabs}
\usepackage{colortbl}
\usepackage{multirow}
\usepackage{hhline}
\usepackage{arydshln}
\usepackage{numprint}
\usepackage[normalem]{ulem}
\usepackage[ruled]{algorithm2e}
\usepackage{wrapfig}

\definecolor{gray-bg}{gray}{0.9}

\usepackage[acronym]{glossaries}
\newacronym{ours}{\mbox{Genifer}}{GENerative Image replay driven by FEatuRes}
\newcommand{\configlft}[3]{\textcolor{darkgray}{(#1+\bm{#2} \times #3)}}
\newcommand{\config}[2]{\textcolor{darkgray}{(\bm{#1}\times#2)}}

\usepackage[capitalize]{cleveref}
\crefname{section}{Sec.}{Secs.}
\Crefname{section}{Section}{Sections}
\Crefname{table}{Table}{Tables}
\crefname{table}{Tab.}{Tabs.}

\ifCLASSOPTIONcompsoc
  \usepackage[nocompress]{cite}
\else
  \usepackage{cite}
\fi

\ifCLASSINFOpdf
\else
\fi

\usepackage{stfloats}

\begin{document}

\title{Generative Feature-driven Image Replay\\ for Continual Learning}

\author{Kevin~Thandiackal,
        Tiziano~Portenier,
        Andrea~Giovannini,
        Maria~Gabrani,
        Orcun~Goksel%
\IEEEcompsocitemizethanks{
\IEEEcompsocthanksitem Kevin Thandiackal, Andrea Giovannini, and Maria Gabrani are with IBM Research Europe, Zurich, Switzerland (email: \{kth,agv, mga\}@zurich.ibm.com).\protect\\
\IEEEcompsocthanksitem Kevin Thandiackal, Tiziano Portenier, and Orcun Goksel are with the Computer-assisted Applications in Medicine group at ETH Zurich, Zurich, Switzerland.\protect\\ 
\IEEEcompsocthanksitem Orcun Goksel is also with the Department of Information Technology, Uppsala University, Uppsala, Sweden (email: orcun.goksel@it.uu.se).
}
}

\markboth{Thandiackal \MakeLowercase{\textit{et al.}}: Generative Feature-driven Image Replay for Continual Learning}%
{Thandiackal \MakeLowercase{\textit{et al.}}: Generative Feature-driven Image Replay for Continual Learning}

\IEEEtitleabstractindextext{%
\begin{abstract}
  Neural networks are prone to catastrophic forgetting when trained incrementally on different tasks.
  Popular incremental learning methods mitigate such forgetting by retaining a subset of previously seen samples and replaying them during the training on subsequent tasks.
  However, this is not always possible, e.g., due to data protection regulations.
  In such restricted scenarios, one can employ generative models to replay either artificial images or hidden features to a classifier.
  In this work, we propose \gls{ours} (\textbf{GEN}erat\textbf{I}ve \textbf{FE}ature-driven image \textbf{R}eplay), where a generative model is trained to replay images that must induce the same hidden features as real samples when they are passed through the classifier.
  Our technique therefore incorporates the benefits of both image and feature replay, i.e.: (1)~unlike conventional image replay, our generative model explicitly learns the distribution of features that are relevant for classification; (2)~in contrast to feature replay, our entire classifier remains trainable; and (3)~we can leverage image-space augmentations, which increase distillation performance while also mitigating overfitting during the training of the generative model.
  We show that \gls{ours} substantially outperforms the previous state of the art for various settings on the CIFAR-100 and CUB-200 datasets.
\end{abstract}

}

\maketitle

\IEEEdisplaynontitleabstractindextext

\IEEEpeerreviewmaketitle

\IEEEraisesectionheading{\section{Introduction}\label{sec:introduction}}

\IEEEPARstart{H}{umans} have the innate ability to continuously learn new tasks while remembering and leveraging prior knowledge.
Similarly, it has been shown that neural networks are able to transfer knowledge from one task to another~\cite{Pan2010ALearning}.
Unlike humans, however, when neural networks are fine-tuned on a new task, they suffer from a phenomenon called \emph{catastrophic forgetting}~\cite{McCloskey1989,Goodfellow2013AnNetworks}.
Neural networks forget how to solve previously learned tasks when trained on subsequent ones.
This very problem lies at the core of Continual Learning (CL), a field that explores techniques to learn different tasks sequentially~\cite{Parisi2018ContinualReview,Farquhar2018TowardsLearning,vandeVen2019ThreeLearning,DeLange2019ATasks}.
In CL scenarios, training data is provided as a stream of tasks, i.e., training data is accessible neither before nor after learning the respective task. Such settings are common constraints in real world applications where data providers cannot grant unlimited and indefinite access to data for various reasons.
Besides acquisition-related factors, legal constraints may exist when processing confidential or privacy-regulated data, such as personal or medical patient data. In this work, we address class-incremental learning (CIL), a specific CL scenario where a classifier is to incrementally learn different classification tasks~\cite{masana2020class,Belouadah2020ATasks}. CIL is a popular experimental setting for analyzing CL techniques, since it enables evaluation on standard benchmark datasets.
For instance, several previous works~\cite{Rebuffi2017ICaRL:Learning,HouLearningRebalancing,LiuMnemonicsForgetting,yu2020semantic,ZhuPASS2021,toldoFusion2022} reported evaluation results with well-defined task splits on CIFAR-100~\cite{Krizhevsky2009LearningImages} and CUB-200~\cite{Wah2011TheDataset}, which are well-suited to stress-test CIL methods in diverse settings.
The former dataset with daily objects demonstrates a scenario with small images and a large range of variation between class appearances, whereas the latter with different bird species contains higher-resolution images with relatively minute differences between classes.

Many successful techniques in CIL preserve previous knowledge by employing some form of \emph{experience replay}, where previously seen samples are replayed during the acquisition of new knowledge~\cite{Rebuffi2017ICaRL:Learning,CastroEnd-to-EndLearning,Wu2019LargeLearning,HouLearningRebalancing,BelouadahScaIL:Learning,LiuMnemonicsForgetting}.
However, as mentioned above, retaining all or even some of the training data is not always possible.
To address this challenge, potential solutions have become feasible thanks to recent advances in generative modeling, in particular generative adversarial networks~(GANs)~\cite{Goodfellow2014GenerativeNets}.
GANs enable so-called \emph{generative replay}, where a generator is trained to replay artificial experience. Leveraging generative models has been shown to be effective in tackling such stricter CIL settings, e.g.\ in~\cite{Shin2017ContinualReplay,CongGANForgetting,OstapenkoLearningLearning,WuMemoryForgetting,vandeVen2020Brain-inspiredNetworks,Kemker2017FearNet:Learning,Liu2020GenerativeLearning,XiangIncrementalNetworks}.

Methods that build on generative replay either perform \emph{image replay} or \emph{feature replay}.
In image replay~\cite{Shin2017ContinualReplay,OstapenkoLearningLearning,WuMemoryForgetting,CongGANForgetting}, a generative model replays artificial samples following the training data distribution, which facilitates knowledge preservation throughout the entire classifier.
However, training such models on complex image datasets such as ImageNet~\cite{Russakovsky2015ImageNetChallenge} is non-trivial.
Recently, an alternative approach of synthesizing artificial hidden features learned by the classifier has shown success for replaying complex data~\cite{Kemker2017FearNet:Learning,vandeVen2020Brain-inspiredNetworks,Liu2020GenerativeLearning,XiangIncrementalNetworks}.
A major advantage of such feature replay is that the distribution to be learned by the generative model is typically significantly simpler and often lower-dimensional than the respective image distribution.
This enables efficient application in real world scenarios beyond moderately complex datasets. In addition, the replayed features are tailored to the knowledge learned by the classifier, which facilitates highly efficient knowledge preservation.
In contrast to image replay, feature replay however cannot leverage standard image-space augmentations, which have been proven to be effective for image classification problems~\cite{perez2017effectiveness}.
Furthermore, feature replay mitigates catastrophic forgetting only in a portion of the classifier, i.e., forgetting in the feature extraction layers is not addressed by design.
Feature-replay techniques thus often freeze the feature extractor~\cite{vandeVen2020Brain-inspiredNetworks,XiangIncrementalNetworks}, which in turn prohibits further knowledge acquisition in these respective layers.

\begin{figure}
    \centering
    \includegraphics[width=\linewidth]{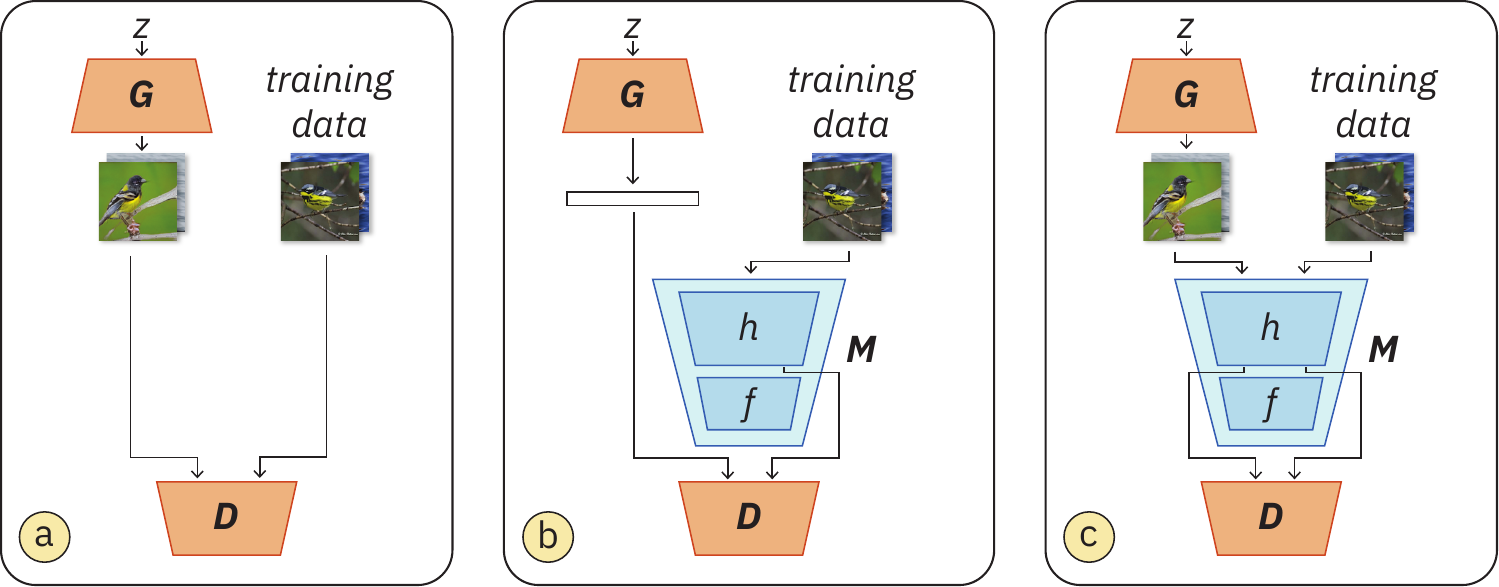}
    \caption{Training of a GAN consisting of $G$ and $D$ to replay images (a), hidden features (b), and our proposed technique (c) for a classifier $M = f \circ h$}
    \label{fig:feature_matching}
\end{figure}

In this work, we propose a novel generative replay technique that combines the advantages of both image and feature replay.
In particular, we train a GAN to synthesize images, similar to generative image replay.
However, our discriminator does not judge the generated images, but the response from the classifier's feature extractor, as in generative feature replay.
Since we replay images, our approach enables preservation and acquisition of knowledge throughout all classifier layers, as the entire classifier remains trainable.
Moreover, image-space augmentations can be leveraged to increase knowledge preservation.
In contrast to naive image replay, our generative model is only tasked with reproducing the distribution of hidden features previously learned by the classifier, as opposed to the underlying image distribution. Our technique thus also inherits the advantages of feature replay, such as scalability to complex training data and classifier-tailored replay.
Therefore, we refer to our technique as \emph{generative feature-driven image replay}.
\cref{fig:feature_matching} shows a visual comparison of training a GAN to replay images, hidden features, and our hybrid approach.

In this work, we make the following contributions:
\begin{enumerate}
    \item A novel generative replay technique that combines the advantages of image and feature replay.
    \item An \emph{adversarial distillation} that improves knowledge preservation in the generative model.
    \item Comprehensive benchmarking on two datasets with different image resolution and class granularity, with varying numbers of tasks, and with and without a large initial task, all showing that our method outperforms the previous state of the art.
\end{enumerate}

\section{Related Work}
Most CL approaches rely on either \emph{regularization}, \emph{isolated parameters}, \emph{replay}, or a combination thereof~\cite{DeLange2019ATasks}.
The main disadvantage of methods based on regularization~\cite{Kirkpatrick2017OvercomingNetworks,Nguyen2018VariationalLearning,Zenke2017ContinualIntelligence} or isolated parameters~\cite{MallyaPiggyback:Weights,Serra2018OvercomingTask} is that they typically require task membership information at inference time~\cite{Farquhar2018TowardsLearning,vandeVen2019ThreeLearning}.
This significantly limits their applicability, since such prior knowledge is not available in most practical applications.
Even with approaches using dedicated task inference techniques~\cite{Wortsman2020SupermasksSuperposition}, the classifier cannot learn any inter-task relationships because it never sees data from different tasks simultaneously.
Approaches that leverage experience replay have been shown to be effective at learning such inter-task relationships~\cite{vandeVen2020Brain-inspiredNetworks,Lesort2020ContinualProcesses,Lesort2021RegularizationLearning}.
For instance, iCaRL~\cite{Rebuffi2017ICaRL:Learning} replays selectively stored training samples, so-called \emph{exemplars}.
Since memory constraints on the number of storable exemplars may induce an imbalance between exemplars and current data, it was shown that addressing such imbalance is essential in this approach~\cite{CastroEnd-to-EndLearning,HouLearningRebalancing,Wu2019LargeLearning,BelouadahScaIL:Learning}.

\subsection{Generative Replay}
Recently, thanks to progress in generative modeling, it has become feasible to address stricter CL settings where no exemplars can be stored.
Furthermore, employing a generative model to replay artificial experience circumvents the above-mentioned imbalance problem, since artificial experience can be sampled from the generative model.
Such promising \emph{generative replay} techniques are particularly relevant for our work:
Deep Generative Replay (DGR)~\cite{Shin2017ContinualReplay} demonstrated that training a GAN is effective for replaying artificial images from MNIST~\cite{lecun1998gradient}.
However, training GANs to generate high-resolution images of complex datasets is challenging and an active research topic on its own.
Therefore, methods such as Always Be Dreaming (ABD)~\cite{SmithABD2021} utilize data-free training for generative models, but this approach relies on a crude approximation of previous task data (batch normalization statistics).
Another direction to alleviate this problem is to replay generated hidden features learned by the classifier, instead of raw images.
Such generative \emph{feature replay} can be highly effective, since the replayed representation is then tailored to the knowledge learned by the classifier.
This approach indeed bares similarities to biological neural networks in the brain, where mental images are not passed back to the retina either~\cite{vandeVen2020Brain-inspiredNetworks}.
Accordingly, in Brain-Inspired Replay (BI-R)~\cite{vandeVen2020Brain-inspiredNetworks}, a variational autoencoder (VAE)~\cite{Kingma2013Auto-EncodingBayes} and a classifier are merged into a single model that internally replays previously learned representations. Similarly, in FearNet~\cite{Kemker2017FearNet:Learning} a mechanism is proposed for generative feature replay combining short- and long-term memory modules.
Unlike these brain-inspired approaches, Liu et al.~\cite{Liu2020GenerativeLearning} employ a conditional GAN~\cite{Mirza2014ConditionalNets} that learns to perform feature replay.

To avoid catastrophic forgetting in the classifier layers that extract the replayed features, previous state-of-the-art generative feature replay methods rely on pretrained, frozen feature extractors~\cite{XiangIncrementalNetworks,vandeVen2020Brain-inspiredNetworks,Kemker2017FearNet:Learning}. 
A drawback of this approach is that it prevents the feature extractor from being trained continually.
In~\cite{Liu2020GenerativeLearning}, a feature distillation loss is proposed as an alternative to freezing.
However, feature distillation, when minimized using \emph{current} samples, only offers a relaxation of the freezing paradigm, without providing fundamental advantages in addition; i.e., the classifier is neither encouraged to learn new features, nor effectively prevented from changing its learned representation of previous data.
Generative Feature Replay with Orthogonal Weight Modification (GFR-OWM)~\cite{ShenGFROWM2021} also employs generative feature replay, but has only been shown to work for very small CNNs, which limits its applicability.
DeepCollab~\cite{CuiDeepCollab2021} is another generative replay method based on a latent feature distribution. 
Since it consists of several customized components (GAN, VAE, domain adaptors, sample selection) and requires multiple training phases per task, its training and tuning are non-trivial.

In contrast to the aforementioned methods, we propose a hybrid approach.
By replaying \emph{images} that represent the distribution of classifier \emph{features}, our technique leverages the benefits of both feature replay and image replay.
As the replayed images are classifier-tailored, our classifier can remember tasks efficiently, while acquiring new knowledge effectively throughout the entire classifier.
Our model can therefore be trained to replay complex and high-dimensional distributions such as CUB-200, which enables successful continual learning.

\subsection{Forgetting in the Generative Model}
Generative replay techniques partially shift the problem of catastrophic forgetting from the classifier to a generative model, since the latter must be trained continually as well.
Therefore, such frameworks must in turn address forgetting in the generative model. 
Wu et al.~\cite{WuMemoryForgetting} proposed a generator distillation loss and demonstrated its effectiveness on moderately complex task sequences. 
However, such image-space distillation leads to increasingly corrupted samples over long task sequences~\cite{CongGANForgetting}. 
Another strategy inspired by isolated parameters is to expand the generative model incrementally using task-specific parameters~\cite{OstapenkoLearningLearning,CongGANForgetting}. 
By employing sparsity regularization~\cite{OstapenkoLearningLearning} or singular value decomposition~\cite{CongGANForgetting} for task-specific parameters, these techniques have been shown to scale to long task sequences.
In our approach, we propose novel adversarial distillation losses (see~\Cref{sec:gen_training}) to address forgetting in the generative model.
Our distillation strategy avoids replaying corrupted samples in long task sequences and at the same time mitigates overfitting in the discriminator.

\subsection{Prototypical Representations}
As an alternative to generative feature replay, recent methods were proposed to employ class-representative prototypes in feature space, i.e., a mean feature for each class.
Prototype Augmentation and Self-Supervision (PASS)~\cite{ZhuPASS2021} stores such prototypes and uses self-supervised learning to learn more general features.
When new classes are learned, the prototypes are augmented with Gaussian noise.
However, there is no guarantee that the real features follow a Gaussian distribution.
In addition to using prototypical representations, Semantic Drift Compensation (SDC)~\cite{yu2020semantic} estimates the drift of previous task prototypes when learning new tasks.
This drift estimation is then used to update the prototypes, leading to superior performance compared to retaining outdated prototypes.
Fusion~\cite{toldoFusion2022}, which is the state of the art in this field, extends SDC by distinguishing semantic and feature drift, where the former models the relationship between new and old class prototypes \emph{before} learning a new task while the latter estimates how the new class representations move \emph{during} the learning of a new task.
Both PASS and Fusion employ feature distillation, but since previous task prototypes are only available in the feature space, feature distillation is performed using current task samples.
As described earlier, this is suboptimal and hinders the learning of new tasks.
Furthermore, image-space augmentations on previous samples can then not be leveraged.
Since we generate previous samples in the image space, our method can both distill features of previous samples and leverage image-space augmentations, leading to results superior to Fusion.

\section{Generative Feature-driven Image Replay}
We herein introduce \gls{ours} (\textbf{GEN}erat\textbf{I}ve \textbf{FE}ature-driven image \textbf{R}eplay), a novel framework for CIL.
It consists of a generative replay model $G$ and a classifier $M$ that are trained in an alternating fashion.
We first describe these components, then discuss the importance of data augmentation in CIL, and finally explain essential implementation details.

In a classical supervised classification problem with i.i.d.\ data, $M$ is trained on tuples $(\mathbf{x}, y)$ sampled from a distribution $\mathcal{S} = \mathcal{X} \times \mathcal{C}$, where $\mathcal{C} = \{1, ..., C\}$ is a set of $C$ classes. In CIL, however, we are presented with a partition of $\mathcal{S}$ consisting of subsets $\mathcal{S}_t = \mathcal{X}_t \times \mathcal{C}_t$, where $t \in \{1, ..., T\}$ is a task, associated to a point in time.
Note that each subset contains samples belonging to a unique set of classes $\mathcal{C}_t$ with $\mathcal{C}_t \cap \mathcal{C}_{t'} = \emptyset$ for $ t \neq t'$.
When learning a task $t$, we denote previously seen task data (i.e., $\mathcal{S}_{1:t-1}$) as $\mathcal{S}_{\mathrm{p}} = \mathcal{X}_{\mathrm{p}} \times \mathcal{C}_{\mathrm{p}}$ and current task data  as $\mathcal{S}_{\mathrm{c}} = \mathcal{X}_{\mathrm{c}} \times \mathcal{C}_{\mathrm{c}}$.

\begin{figure*}
    \centering
    \includegraphics[width=\linewidth]{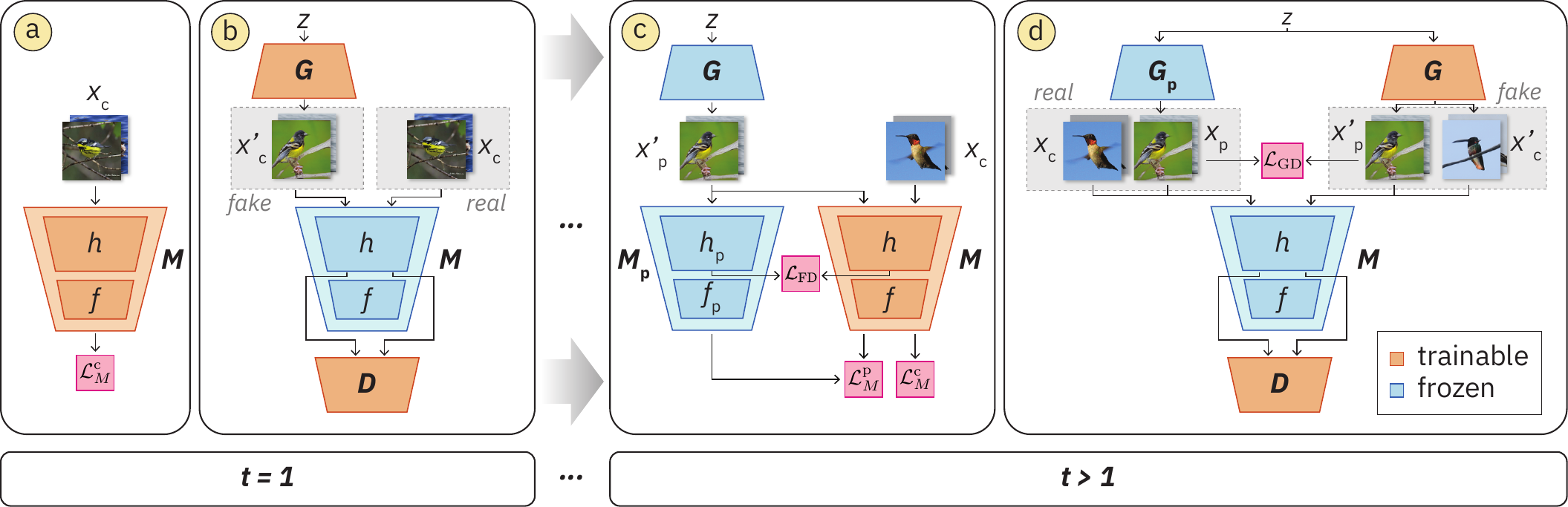}
    \caption{Training procedure for our method.
    First, classifier $M$~(a) and then $(G, D)$~(b) are trained on task $t = 1$.
    For any future task $t > 1$, classifier $M$~(c) and then $(G, D)$~(d) learn the current task, while minimizing additional losses proposed to prevent forgetting on previous tasks. Note that at a task $t$, only the latest GAN is used, i.e., no previous GANs need to be retained.}
    \label{fig:overview}
\end{figure*}

During training for new tasks, our method alternates between training a classifier and a GAN.
This is illustrated in~\cref{fig:overview} to aid the descriptions in the following subsections.

\subsection{Generative Replay Model} \label{sec:gen_training}
As generative model, we train a conditional GAN consisting of a generator $G$ and a projection discriminator $D$~\cite{Miyato2018CGANsDiscriminator}.
Given a class $y$ and a randomly sampled latent vector~$\mathbf{z}$, $G$$:$$(\mathbf{z}, y)$$\mapsto$$\mathbf{x'}$ learns to create synthetic samples $\mathbf{x'}$ belonging to class~$y$.
Since $G$ is trained in a continual fashion, it must remember how to generate previous-task samples while learning to synthesize new samples representing the current task. 
We achieve this by minimizing the objectives $\mathcal{L}_{G} = \mathcal{L}_{G}^{\mathrm{c}} + \mathcal{L}_{G}^{\mathrm{p}}$ and $\mathcal{L}_{D} = \mathcal{L}_{D}^{\mathrm{c}} + \mathcal{L}_{D}^{\mathrm{p}}$, respectively,  for $G$ and $D$. 
Losses $\mathcal{L}_{G}^{\mathrm{c}}$ and $\mathcal{L}_{D}^{\mathrm{c}}$ are computed on samples of the current task so that $G$ learns to synthesize realistic new-task samples.
$\mathcal{L}_{G}^{\mathrm{p}}$ and $\mathcal{L}_{D}^{\mathrm{p}}$ constitute \emph{adversarial distillation} losses, which are computed on samples belonging to previous tasks.
Their purpose is to prevent $G$ from forgetting previously learned samples.

We define the classifier $M$$:$$\mathbf{x}$$\mapsto$$f\left(h\left( \mathbf{x} \right)\right)$, where $h$ is a function that extracts features up to a certain convolutional layer in $M$, and $f$ (potentially consisting of additional convolutional layers) maps this representation to output logits for each class. $\mathcal{L}_{G}^{\mathrm{c}}$ is the non-saturating logistic loss~\cite{Goodfellow2014GenerativeNets}
\begin{equation}
    \mathcal{L}_{G}^{\mathrm{c}} = \mathop{\mathbb{E}}_{\substack{\mathbf{z} \sim p_z \\ y \sim p_{\mathcal{C}_{\mathrm{c}}}}} \bigg[ a \Big(-D\big(h\big(G(\mathbf{z}, y)\big), y\big)\Big) \bigg], \label{eq:l_gn_curr}
\end{equation}
where $a$ is the softplus function.
Note that although $G$ synthesizes images $\mathbf{x'}$, $D$ discriminates their representation ($h(\mathbf{x'})$) learned by $M$. This is a crucial difference from all previous techniques, where $D$ would discriminate $G(\mathbf{z}, y)$ directly. 
Our replayed images thus contain the desired features for optimal classifier distillation without being constrained in any other way, which greatly simplifies the training of the generative model.

The adversarial distillation loss for $G$ is minimized starting from the second task onward and is defined as
\begin{equation}
    \begin{aligned}
    \mathcal{L}_{G}^\mathrm{p} =
        & \overbrace{\mathop{\mathbb{E}}_{\substack{\mathbf{z} \sim p_z \\ y \sim p_{\mathcal{C}_\mathrm{p}}}} \bigg[ a \Big(-D \big( h \big( G(\mathbf{z}, y) \big), y\big) \Big) \bigg]}^{\mathcal{L}_{G^\mathrm{ADV}}^\mathrm{p}} \\
        &+ \lambda_\mathrm{ID} \underbrace{\mathop{\mathbb{E}}_{\substack{\mathbf{z} \sim p_z \\ y \sim p_{\mathcal{C}_\mathrm{p}}}} \bigg[ \left\lVert G(\mathbf{z}, y) - G_\mathrm{p}(\mathbf{z}, y) \right\rVert_1 \bigg]}_{\mathcal{L}_{G^\mathrm{ID}}^\mathrm{p}} \;,
    \end{aligned}
    \label{eq:l_gn_prev}
\end{equation}
where $G_\mathrm{p}$ is a frozen copy of the generator from the previous task.
The first component $\mathcal{L}_{G^\mathrm{ADV}}^\mathrm{p}$ is an adversarial loss that encourages $G$ to generate previous task images containing realistic features, similarly to the current task objective in~\cref{eq:l_gn_curr}. 
The second component $\mathcal{L}_{G^\mathrm{ID}}^\mathrm{p}$ is an $\ell_1$ is a loss for image distillation (ID) evaluated in the image space, explicitly penalizing $G$ when it deviates from previously generated samples. 
It is weighted by a hyperparameter $\lambda_\mathrm{ID}$.
Our adversarial distillation differs markedly from previous works (e.g.,~\cite{Liu2020GenerativeLearning}), where only ID was employed. 
Minimizing $\mathcal{L}_{G}^\mathrm{p}$ mitigates the degradation of image quality over a task sequence (e.g., blurring), and we show its superiority compared to previous approaches later in~\cref{sec:abl_dist}.

Analogously to the generator loss, the objective for $D$ consists of two components: $\mathcal{L}_{D}^\mathrm{c}$ for learning the current task and $\mathcal{L}_{D}^\mathrm{p}$ for the adversarial distillation on previous task samples. With the first component
\begin{equation}
    \mathcal{L}_{D}^\mathrm{c}\!=\!\!\!\!
        \mathop{\mathbb{E}}_{\substack{\mathbf{z} \sim p_z \\ y \sim p_{\mathcal{C}_\mathrm{c}}}}\!\! \bigg[ a \Big(\!D \big(h \big( G(\mathbf{z}, y) \big), y\big) \Big) \bigg]
        +\!\!\mathop{\mathbb{E}}_{\mathbf{x},y \sim p_{\mathcal{S}_\mathrm{c}}}\!\! \bigg[a \Big(\!-D \big(h(\mathbf{x}), y\big) \Big)\bigg],
    \label{eq:l_dn_curr}
\end{equation}
$D$ learns to distinguish between features $h(\mathbf{x})$ from real images and features $h(G(\mathbf{z}, y))$ from generated images, thereby teaching $G$ to synthesize images with features that look realistic from the classifier's perspective.
For the second component $\mathcal{L}_{D}^\mathrm{p}$, we do not have access to real previous task samples ($\mathcal{S}_\mathrm{p}$).
We thus use $G_\mathrm{p}$ to \emph{generate} artificial samples $\mathcal{S}'_\mathrm{p}$ that are representative of the previous tasks. 
The second component is then defined as
\begin{equation}
    \mathcal{L}_{D}^\mathrm{p}\!=\!\!\!\!
        \mathop{\mathbb{E}}_{\substack{\mathbf{z} \sim p_z \\ y \sim p_{\mathcal{C}_\mathrm{p}}}} \!\! \bigg[ a \Big( D \big( h \big( G(\mathbf{z}, y) \big), y\big) \Big)
        + a \Big(\!\!-\!D\big(h \big( G_\mathrm{p}(\mathbf{z}, y) \big), y\big) \Big)\bigg].
    \label{eq:l_dn_prev}
\end{equation}
This loss helps $G$ to remember how to generate realistic previously-learned samples.
As before, $\mathcal{L}_{D}^\mathrm{p}$ is only minimized by $D$ starting from the second task onwards. 
Thanks to the adversarial distillation in $D$, the risk of discriminator overfitting is reduced significantly, since $D$ is trained on $\mathcal{S}_\mathrm{c} \cup \mathcal{S}'_\mathrm{p}$, which is substantially larger and more diverse than $\mathcal{S}_\mathrm{c}$.
Note that this technique implies a label flip for $D$:
once a task is learned and it becomes a ``previous'' task, the labels of the corresponding samples are changed from fake to real.
Finally, $D$ also minimizes an $R_1$ regularization term~\cite{Mescheder2018WhichConverge} defined as $\lambda_{R_1}\mathop{\mathbb{E}}{[ \lVert \nabla_{h(\mathbf{x})} D(h(\mathbf{x})) \rVert^2_2 ]}$.

\subsection{Classifier Training} \label{sec:clsf_training}
To learn a new task consisting of $L$ classes, the function $f$ in $M$ is extended with weights for the additional $L$ output logits. 
$M$ then minimizes the cross-entropy loss $\mathcal{L}_M^\mathrm{c}$ on current real samples.
At this point in time, we also have access to a copy of the previous classifier $M_\mathrm{p}$ and a generator $G$, both of which have been trained on the $K$ classes of $\mathcal{S}_\mathrm{p}$.
For a given image $\mathbf{x} \in \mathcal{X}_\mathrm{c}$, we define the $k$-th logit of $M$ as $M^k(\mathbf{x})$. Analogously, for a synthetic image $\mathbf{x'} = G(\mathbf{z},y)$ (belonging to a previously learned task), the $k$-th logit of $M_\mathrm{p}$ is denoted by $M_{\mathrm{p}}^k(\mathbf{x'})$. The corresponding predicted class probabilities are then defined as:
\begin{equation}
    q^{k}(\mathbf{x}) = \frac{e^{M^k(\mathbf{x})}}{\sum_{j=1}^{K+L}e^{M^{j}(\mathbf{x})}},\ \ \
    q_{\mathrm{p}}^k(\mathbf{x'}) = \frac{e^{M_{\mathrm{p}}^k(\mathbf{x'})}}{\sum_{j=1}^{K}e^{M_{\mathrm{p}}^j(\mathbf{x'})}} \;.
\end{equation}
To mitigate forgetting previous tasks, $M$ minimizes a knowledge distillation loss~\cite{Hinton2015DistillingNetwork}.
For a synthetic previous image $\mathbf{x'} = G(\mathbf{z},y)$, we define our logit distillation (LD) loss as
\begin{equation}
    \mathcal{L}_{M}^\mathrm{p} = - \sum_{k=1}^{K}{q_{\mathrm{p}}^k(\mathbf{x'}) \log(q^{k}(\mathbf{x'}))} \;.
\end{equation}

We propose the feature distillation loss
\begin{equation}
    \mathcal{L}_\mathrm{FD} = \left\lVert h(\mathbf{x'}) - h_\mathrm{p}(\mathbf{x'})\right\rVert^2_2 \;,
\end{equation}
which, in contrast to earlier works~\cite{Liu2020GenerativeLearning,ZhuPASS2021,toldoFusion2022}, is evaluated on synthesized previous task images $\mathbf{x'}$, not current ones.
This is highly effective as shown later in~\cref{sec:abl_dist}.
The final training objective for $M$ is thus 
\begin{equation}
\mathcal{L}_{M} = (1-\lambda_\mathrm{LD})\mathcal{L}_{M}^\mathrm{c} + \lambda_\mathrm{LD} \; \mathcal{L}_{M}^\mathrm{p} + \lambda_\mathrm{FD} \; \mathcal{L}_\mathrm{FD} \; ,
\end{equation}
where $\lambda_\mathrm{LD}$ and $\lambda_\mathrm{FD}$ are the logit and feature distillation coefficients, respectively.

\subsection{Leveraging Augmentations} \label{sec:leveraging_aug}
When training large networks on small datasets, there is a risk of overfitting to training data, which results in poor generalization performance.
This applies not only to classifiers, but also to discriminators in GANs.
In fact, multiple recent works identified discriminator overfitting as a root cause for inferior GAN performance on limited training data~\cite{Karras2020TrainingData,Zhao2020ImageTraining,Tran2020OnTraining}.
As a remedy, Karras et al.~\cite{Karras2020TrainingData} proposed an effective pipeline for adaptive discriminator augmentation (ADA).
Since the dataset sizes per task in CIL are typically small compared to the vast amount of images GANs are usually trained on in non-CIL scenarios (e.g., a task from a long CIL sequence in CUB-200 contains only 150 training images), discriminator overfitting is indeed particularly relevant in CIL.
To the best of our knowledge, discriminator overfitting has neither been addressed nor investigated in the context of generative replay for CIL.

In our framework, we tackle the problem of discriminator overfitting from two orthogonal directions.
First, since we replay images, we augment generated and real images using the augmentations proposed in~\cite{Karras2020TrainingData}.
Note that this would not be possible in the case of feature replay, since respective image space augmentations are not applicable in feature space.
Second, our discriminator is not only trained on current real data but also on synthesized previous data (adversarial distillation in \cref{eq:l_dn_prev}).
We find this to be crucial to effectively mitigate discriminator overfitting on the current task, and we show this to greatly improve the results in~\cref{sec:abl_dist}.

In addition to discriminator overfitting, we can also address overfitting in the classifier.
Specifically, our proposed approach enables us to leverage common image space augmentations (e.g., random cropping and flipping) on replayed samples for logit and feature distillation. Again, note that in the case of feature replay, such classifier augmentations on replayed samples cannot be employed.
In~\cref{sec:abl_augmentations}, we show that both discriminator and classifier augmentations are essential in reducing forgetting over long task sequences.

\subsection{Implementation Details} \label{sec:implementation}
A pseudocode summarizing our training process is given in~\cref{alg:GENIFER}.
    \begin{algorithm}
        \footnotesize
        \caption{Pseudocode for training Genifer} \label{alg:GENIFER}
        \SetKwInOut{Input}{Input}
        \SetKwInOut{Output}{Output}
        
        \Input{Sequence of training data $\mathcal{S}_1, ..., \mathcal{S}_T$; $M, G, D$}
        \Output{$M$ trained on $\mathcal{S}_1, ..., \mathcal{S}_T$}
        Train $M$ to minimize $\mathcal{L}_{M}^\mathrm{c}$\;
        Train $G, D$ to minimize $\mathcal{L}_{G}^\mathrm{c},  \mathcal{L}_{D}^\mathrm{c}$\;
        \For{$t=2, ..., T$}{
            $M_\mathrm{p} \gets M$\;
            Train $M$ to minimize $\mathcal{L}_{M}$\;
            \If{$t \neq T$}{
                $G_\mathrm{p} \gets G$\;
                Train $G, D$ to minimize $\mathcal{L}_{G}, \mathcal{L}_{D}$\;
            }
        }
        \Return $M$\;
    \end{algorithm}
We train a ResNet-18 classifier~\cite{He2016DeepRecognition} using the RAdam optimizer~\cite{Liu2020OnBeyond} with a learning rate of $0.0001$ and a weight decay of $0.0005$.
We empirically found that the best location to divide the classifier $M$ into $h$ and $f$ is after the third ResNet block (i.e., roughly at 3/4 of the classifier depth), which we used throughout all our experiments.
Between experimental settings with varying characteristics (dataset nature, number of images, image size, number of classes, etc.), we tuned the logit distillation weight in the range $\lambda_\mathrm{LD}$$\in$$[0.95,0.99]$ per each setting.
For the setting in Tab.2 without a large first task, we used $\lambda_\mathrm{LD}$$=$$0.8$, which empirically worked better.

Thanks to our approach, we are able to use relatively low-capacity architectures for $G$ and $D$, as opposed to conventional generative image replay methods~\cite{CongGANForgetting}.
Note that for RGB images of size 128$\times$128, the memory footprint of our GAN ($G, D$) is $\approx$70MB, whereas a memory buffer for exemplar-based methods would typically store 2000 samples occupying $\approx$100MB.
$D$ employs a mini-batch discrimination layer~\cite{karras2017progressive} to increase sample diversity.
Both $G$ and $D$ are trained with an equalized learning rate~\cite{karras2017progressive} of $0.0025$.
The $R_1$ regularization weight is set to $\lambda_{R_1}$$=$$0.5$.
At inference time (i.e., when replaying from $G$ at test time or during training to synthesize previous-task samples), we use a copy of $G$ with exponential moving average over the training iterations applied to its parameters~\cite{Yazc2019TheTraining}.
For image distillation, we set $\lambda_\mathrm{ID}$$=$$10 \kappa$, where $\kappa$$=$$\frac{|\mathcal{C}_\mathrm{p}|}{|\mathcal{C}_\mathrm{c}|}$ is the ratio between the number of previous and current classes.
Through this scaling we increase the weight of the generator distillation as more tasks are added.
We noticed that in order to mitigate generator forgetting over various task sequence lengths, this approach is easier to tweak than a constant distillation coefficient.

All models were implemented in PyTorch~\cite{Paszke2019PyTorch:Library}
and trained using a single NVIDIA A100 Tensor Core GPU.
In our experiments, training the classifier and the GAN for one task takes between one and four hours, depending on the task size and the dataset.
Please refer to the supplemental material for more details and visualizations.

\section{Experiments} \label{sec:exp}

We evaluate our method on two popular CIL benchmark datasets, CIFAR-100~\cite{Krizhevsky2009LearningImages} and CUB-200~\cite{Wah2011TheDataset}.
Both datasets have a large number of classes, which allows us to simulate diverse CIL settings with different task sequence lengths.
In addition, CIFAR-100 and CUB-200 have very dissimilar properties and inherent difficulties, making them ideal to test the general applicability of different CIL methods.
While the major challenge in CIFAR-100 is the diversity among its classes, the main difficulty in CUB-200 comes from the high-resolution images, the requirement to distinguish a large number of fine-grained classes, and the low number of available training images.
Furthermore, both datasets have been commonly used in related previous works~\cite{Rebuffi2017ICaRL:Learning,HouLearningRebalancing,yu2020semantic,toldoFusion2022}, which enables a fair comparison of our method.
CIFAR-100 comprises images of size $32$$\times$$32$ belonging to 100 different object categories, split into \numprint{50000} training and \numprint{10000} test samples.
CUB-200 consists of \numprint{5994} training and \numprint{5794} test images grouped into 200 classes of different birds, which amounts to at most 30 training samples per class. 
CUB-200 images are of variable resolution, with the majority being roughly of size $500 \times 500$.
In line with previous works~\cite{yu2020semantic,toldoFusion2022}, we train our ResNet-18 classifier from scratch for CIFAR-100 and start with a pretrained backbone for CUB-200.
On CIFAR-100, the classifier is trained on each incremental task for 100 epochs with $\lambda_\mathrm{FD}$$=$$1$ while the learning rate decays with factor 5 after 30, 60, and 80 epochs.
Similarly on CUB-200, the classifier is trained for 200 epochs with $\lambda_\mathrm{FD}$$=$$0.1$ while the learning rate decays after 60, 120, and 160 epochs.

To evaluate incremental learning performance, the datasets must be split into a sequence of tasks.
Most related previous works~\cite{HouLearningRebalancing,Liu2020GenerativeLearning,ZhuPASS2021,yu2020semantic,toldoFusion2022} have adopted a standardized experimental setup, which we follow here to enable a direct comparison.
In this standard setting, the first task $t$$=$$1$ is larger and consists of learning half of the classes in the dataset; except for one setting in CIFAR-100, where the first task contains 40 classes.
The remaining dataset is then uniformly split into 5, 10, or 20 tasks.
This setting indeed reflects typical CIL use cases, where an initial model would be trained only when a sufficiently large dataset has been collected and is available.
For completeness, we additionally show a comparison on CIFAR-100 without a larger first task, i.e., all the tasks (including $t=1$) comprise the same number of classes.
Note that most previously proposed methods only specialize and provide results for one scenario, either with or without large first task; which places our paper uniquely, presenting results from both scenarios.

\subsection{Comparisons with the state of the art} \label{sec:res}
\begin{table*}[t]
\caption{Average incremental accuracy $\alpha$ and overall accuracy $\alpha_T$ (\%) on CIFAR-100 and CUB-200 for different numbers of tasks. The best and second-best results are shown in \textbf{bold} and \underline{underlined}, respectively.}
    \centering
    \begin{tabular}{clcccccccccccc}
    \toprule
    & Dataset & \multicolumn{6}{c}{CIFAR-100} & \multicolumn{6}{c}{CUB-200} \\
    \cmidrule(lr){3-8}\cmidrule(lr){9-14}
    & Tasks & \multicolumn{2}{c}{$\underset{\configlft{50}{5}{10}}{\mathbf{5}}$} & \multicolumn{2}{c}{$\underset{\configlft{50}{10}{5}}{\mathbf{10}}$} & \multicolumn{2}{c}{$\underset{\configlft{40}{20}{3}}{\mathbf{20}}$} & \multicolumn{2}{c}{$\underset{\configlft{100}{5}{20}}{\mathbf{5}}$} & \multicolumn{2}{c}{$\underset{\configlft{100}{10}{10}}{\mathbf{10}}$} & \multicolumn{2}{c}{$\underset{\configlft{100}{20}{5}}{\mathbf{20}}$} \\
    \cmidrule(lr){3-4}\cmidrule(lr){5-6}\cmidrule(lr){7-8}\cmidrule(lr){9-10}\cmidrule(lr){11-12}\cmidrule(lr){13-14}
    & Metric  & \cellcolor{gray-bg}{$\alpha$} & $\alpha_T$ 
    & \cellcolor{gray-bg}{$\alpha$} & $\alpha_T$ 
    & \cellcolor{gray-bg}{$\alpha$} & $\alpha_T$ 
    & \cellcolor{gray-bg}{$\alpha$} & $\alpha_T$
    & \cellcolor{gray-bg}{$\alpha$} & $\alpha_T$
    & \cellcolor{gray-bg}{$\alpha$} & $\alpha_T$ \\
    \midrule
    & Lower bound & \cellcolor{gray-bg}{23.4} & 9.0 & \cellcolor{gray-bg}{13.2} & 4.7 & \cellcolor{gray-bg}{8.1} & 2.9 & \cellcolor{gray-bg}{25.4} & 9.0 & \cellcolor{gray-bg}{18.3} & 5.1 & \cellcolor{gray-bg}{14.4} & 3.1 \\
    \cmidrule(r){1-2}\cmidrule(lr){3-8}\cmidrule(l){9-14}
    \multirow{2}{*}{EB}
    & iCaRL~\cite{Rebuffi2017ICaRL:Learning} & \cellcolor{gray-bg}{60.7} & 54.8  & \cellcolor{gray-bg}{59.8} & 53.3 & \cellcolor{gray-bg}{55.4} & 47.6 & \cellcolor{gray-bg}{63.2} & 57.9 & \cellcolor{gray-bg}{\underline{62.7}} & \underline{55.7} & \cellcolor{gray-bg}{\underline{58.6}} & \underline{46.9} \\
    & LUCIR~\cite{HouLearningRebalancing} & \cellcolor{gray-bg}{63.3} & 53.7  & \cellcolor{gray-bg}{57.1} & 47.4 & \cellcolor{gray-bg}{53.5} & 44.1 & \cellcolor{gray-bg}{\underline{65.6}} & \underline{60.0} & \cellcolor{gray-bg}{62.2} & 53.0 & \cellcolor{gray-bg}{56.7} & 43.6 \\
    \cmidrule(r){1-2}\cmidrule(lr){3-8}\cmidrule(l){9-14}
    \multirow{4}{*}{EF}
    & PASS~\cite{ZhuPASS2021} & \cellcolor{gray-bg}{65.1} & 56.5 & \cellcolor{gray-bg}{60.8} & 47.6 & \cellcolor{gray-bg}{58.7} & 47.3 & \cellcolor{gray-bg}{60.1} & 52.1 & \cellcolor{gray-bg}{53.3} & 38.0 & \cellcolor{gray-bg}{34.3} & 18.3 \\
    & SDC~\cite{yu2020semantic} & \cellcolor{gray-bg}{66.2} & 57.6 & \cellcolor{gray-bg}{62.7} & 52.3 & \cellcolor{gray-bg}{59.2} & 48.8 & \cellcolor{gray-bg}{60.3} & 52.3 & \cellcolor{gray-bg}{51.6} & 38.3 & \cellcolor{gray-bg}{35.1} & 18.2 \\
    & Fusion~\cite{toldoFusion2022} & \cellcolor{gray-bg}{\underline{66.8}} & \underline{58.7} & \cellcolor{gray-bg}{\underline{65.1}} & \underline{56.9} & \cellcolor{gray-bg}{\underline{61.6}} & \textbf{51.8} & \cellcolor{gray-bg}{63.2} & 57.0 & \cellcolor{gray-bg}{60.0} & 52.6 & \cellcolor{gray-bg}{53.6} & 38.3 \\
    & \gls{ours} (ours) & \cellcolor{gray-bg}{\textbf{69.4}} & \textbf{61.5} & \cellcolor{gray-bg}{\textbf{68.4}} & \textbf{58.9} & \cellcolor{gray-bg}{\textbf{65.0}} & \underline{51.4} & \cellcolor{gray-bg}{\textbf{70.7}} & \textbf{63.0} & \cellcolor{gray-bg}{\textbf{70.3}} & \textbf{61.2} & \cellcolor{gray-bg}{\textbf{66.6}} & \textbf{52.3} \\
    \cmidrule(r){1-2}\cmidrule(lr){3-8}\cmidrule(l){9-14}
    & Upper bound & \multicolumn{6}{c}{76.3} & \multicolumn{6}{c}{75.4} \\
    \bottomrule
    \end{tabular}
\label{tab:summary_lft}
\end{table*}
We evaluate our method using two popular metrics: the overall test accuracy $\alpha_T$ at the end of a sequence containing $T$ tasks, i.e., the final accuracy over all classes in the dataset, and the average incremental accuracy $\alpha$$=$$\frac{1}{T} \sum_{t=1}^{T} \alpha_t$ as introduced in~\cite{Rebuffi2017ICaRL:Learning}.
In \cref{tab:summary_lft} we show results on CIFAR-100 and CUB-200 for settings with a large first task and 5/10/20 subsequent tasks.
We compare our proposed method Genifer with various existing CIL methods; in particular Prototype Augmentation and Self-Supervision (PASS)~\cite{ZhuPASS2021}, Semantic Drift Compensation (SDC)~\cite{yu2020semantic}, and Fusion~\cite{toldoFusion2022} as exemplar-free (EF) approaches with state-of-the-art results on the evaluated datasets.
Their results were taken directly as they were reported in~\cite{toldoFusion2022} using a ResNet-18 as classifier.
Exemplar-based (EB) methods are not direct competitors for our method, as they deal with a simplified scenario and our problem definition focuses on settings where no exemplar can be kept.
Nonetheless, we include herein two popular exemplar-based methods for comparison:
Incremental Classifier and Representation Learning (iCaRL)~\cite{Rebuffi2017ICaRL:Learning} and Learning a Unified Classifier Incrementally via Rebalancing (LUCIR)~\cite{HouLearningRebalancing}.
We based their implementations on the FACIL framework~\cite{masana2020class} using a ResNet-18 backbone.
For their parametrization, following~\cite{HouLearningRebalancing}, a growing exemplar memory buffer is used (20 exemplars per class for CIFAR-100 and 10 exemplars per class for CUB-200), which amounts to 2000 stored exemplars at the end of a sequence.
As reference, we also provide lower bound and upper bound results obtained, respectively, by naive finetuning and joint training of the classifier on the entire dataset.

The results in~\cref{tab:summary_lft} show that our approach outperforms all other methods, including exemplar-based approaches, on both datasets, for 5 out of 6 task settings; while obtaining comparable overall accuracy to Fusion in the 20-task setting on CIFAR-100.
For CIFAR-100, Genifer improves the average incremental accuracy of the previous state of the art (Fusion) by 2.6\,pp (percentage points), 3.3\,pp, and 3.4\,pp in the 5-, 10-, and 20-task settings, respectively.
The improvements are larger on CUB-200, namely 7.5\,pp, 10.3\,pp, and 13.0\,pp.
Note that our dramatic improvements on CUB-200 reduce the gap in overall accuracy $\alpha_T$ between the state of the art and the upper bound by more than 30\,\%, strikingly for each task setting.
Further note that our method is superior even to exemplar-based methods across all task settings.
Indeed in CUB-200, to the best of our knowledge, Genifer is the first exemplar-free CIL method to outperform existing exemplar-based methods.
\begin{table}
\caption{Average incremental accuracy $\alpha$ and overall accuracy $\alpha_T$ (\%) on CIFAR-100 without a large first task. The best and second-best results are shown in \textbf{bold} and \underline{underlined}, respectively.}
    \centering
    \begin{tabular}{llcc}
    \toprule
    & Tasks & \multicolumn{2}{c}{$\underset{\config{5}{20}}{\mathbf{5}}$} \\
    \cmidrule(lr){3-4}
    & Metric  & \cellcolor{gray-bg}{$\alpha$} & $\alpha_T$ \\
    \midrule
    & Lower bound & \cellcolor{gray-bg}{39.5} & 17.3  \\
    \midrule
    \multirow{2}{*}{EB}
    & iCaRL~\cite{Rebuffi2017ICaRL:Learning}      & \cellcolor{gray-bg}{66.3} & 56.2  \\
    & LUCIR~\cite{HouLearningRebalancing}      & \cellcolor{gray-bg}{66.5} & 52.9  \\
    \midrule
    \multirow{5}{*}{EF}
    & GFR-OWM~\cite{ShenGFROWM2021}               & \cellcolor{gray-bg}{48.7} & 36.3 \\
    & ABD~\cite{SmithABD2021}                     & \cellcolor{gray-bg}{-} & 43.9 \\
    & EFT~\cite{verma2021eft}                     & \cellcolor{gray-bg}{65.8} & 52.8 \\
    & DeepCollab~\cite{CuiDeepCollab2021}         & \cellcolor{gray-bg}{\underline{68.6}} & \underline{56.9} \\
    & \gls{ours} (ours)                           & \cellcolor{gray-bg}{\textbf{70.9}} & \textbf{60.0} \\
    \midrule
    & Upper bound & \multicolumn{2}{c}{76.3}\\
    \bottomrule
    \end{tabular}
\label{tab:cifar_no_pre_no_lft}
\end{table}
In~\cref{tab:cifar_no_pre_no_lft} we report results on CIFAR-100 in an experimental setting without a large first task.
Exemplar-free methods specializing in this setting are Generative Feature Replay with Orthogonal Weight Modification (GFR-OWM)~\cite{ShenGFROWM2021}, Always Be Dreaming (ABD)~\cite{SmithABD2021}, Efficient Feature Transformation (EFT)~\cite{verma2021eft}, and DeepCollab~\cite{CuiDeepCollab2021}.
For these, we report here the results presented in their respective papers.
As can be seen in~\cref{tab:cifar_no_pre_no_lft}, we achieve new state-of-the-art performance also in this setting; where
Genifer outperforms the closest method (DeepCollab) in average incremental accuracy and overall accuracy by 2.3\,pp and 3.1\,pp, respectively.

\subsection{Discussion} 
\label{sec:discussion}
Genifer's superior results in~\cref{tab:cifar_no_pre_no_lft} show that our approach excels also in scenarios where it is not feasible to collect a large dataset before starting continual learning.
Furthermore, despite DeepCollab and ABD being based on a deeper ResNet-32 architecture, which is more expressive than our ResNet-18, Genifer still performs better.
This highlights the efficacy of our technique in mitigating catastrophic forgetting.

As seen in~\cref{tab:summary_lft}, our approach achieves the largest gains compared to the previous state of the art on CUB-200.
One potential reason for this may be that SDC~\cite{yu2020semantic} and Fusion~\cite{toldoFusion2022} are mainly based on feature drift estimation and compensation.
In fine-grained datasets such as CUB-200, class clusters may lie close to each other in the feature space, which makes their separation challenging.
This, in turn, can cause slight inaccuracies in feature estimation to become increasingly problematic when learning new tasks while remembering old ones.
In the following, we analyze the effectiveness of our proposed contributions by ablating each major component individually.
We conduct these experiments on CUB-200 since its challenging nature allows us to better highlight differences in performance.

\subsubsection{Feature-driven Image Replay} \label{sec:implicit_feat_replay}
We compare our proposed technique to two obvious alternatives: feature replay (FR) and conventional image replay (IR).
For the first alternative, we train the classifier with FR (as shown in \cref{fig:feature_matching}b), similarly to~\cite{ShenGFROWM2021,vandeVen2020Brain-inspiredNetworks,Liu2020GenerativeLearning}.
To this end, we employ the same discriminator as in our proposed method, but with a modified $G$ to synthesize features instead of images, while maintaining the overall generator capacity.
As in previous work~\cite{Liu2020GenerativeLearning,ZhuPASS2021,toldoFusion2022}, we employ feature distillation with current task samples (as previous task samples are not available in the image space), since we also observed that not doing so performs worse.
Note that, without actual images being synthesized, neither discriminator augmentation nor classifier augmentation can be applied in this scenario.
For the second alternative (IR), similarly to~\cite{WuMemoryForgetting,OstapenkoLearningLearning,CongGANForgetting}, we modify our framework to perform IR, as depicted in \cref{fig:feature_matching}a.
For this purpose, we adapt $D$ to discriminate images instead of features, again without altering its overall capacity. In contrast to FR, data augmentation is applied here the same way as in our proposed framework.
\begin{figure}[t]
     \centering
     \includegraphics[width=0.8\linewidth]{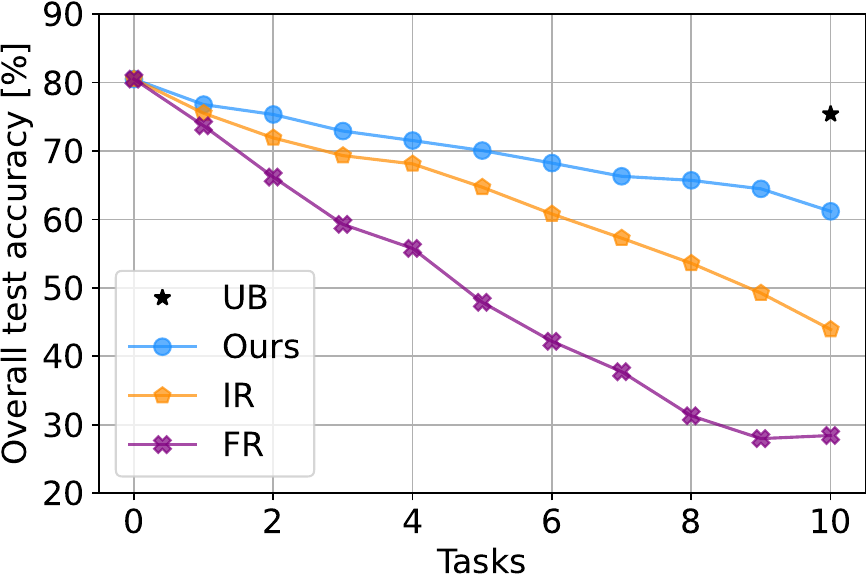}
     \caption{Test accuracy $\alpha_t$ for the upper bound (UB), our proposed method, image replay (IR), and feature replay (FR) on CUB-200 with 10 tasks.}
    \label{fig:ablations}
\end{figure}
To visualize the differences over a task sequence, we show the evolution of $\alpha_t$ over time in \cref{fig:ablations}. 
While IR and our approach are relatively close in accuracy at the beginning of the sequence, their discrepancy steadily increases as more tasks are learned, and our technique clearly outperforms IR at the end of the sequence.
This confirms the hypothesis that in the long run, learning the distribution of features is more effective than learning the distribution of respective images.
FR on the other hand suffers from catastrophic forgetting already early in the sequence, hence performing by far the worst.
One reason for this lower performance is that in FR, feature distillation is performed using current samples, which hinders the model from learning to extract novel features.
The classifier $M$ thus has little capacity left in the remaining layers to learn new tasks without forgetting previous tasks.
Another reason is that data augmentation is not feasible, neither during generator training nor during logit distillation.

\begin{figure}[t]
     \centering
     \begin{subfigure}{0.3\linewidth}
         \centering
         \includegraphics[height=4.5cm]{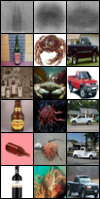}
         \caption{Real}
         \label{fig:var_real}
     \end{subfigure}
     \begin{subfigure}{0.3\linewidth}
         \centering
         \includegraphics[height=4.5cm]{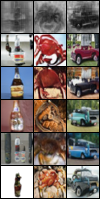}
         \caption{IR}
         \label{fig:var_im}
     \end{subfigure}
     \begin{subfigure}{0.3\linewidth}
         \centering
         \includegraphics[height=4.5cm]{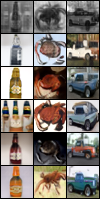}
         \caption{Genifer}
         \label{fig:var_ifm}
     \end{subfigure}
     \caption{Examples of CIFAR-100 images (with columns showing the classes \emph{bottle}, \emph{crab}, and \emph{pick-up truck}) sampled from (a)~the real training dataset, (b)~a generator trained with IR, and (c)~our proposed generator from Genifer. 
    The top row shows the per-pixel variance over all images of the same class.}
    \label{fig:var_imgs}
\end{figure}

To shed more light on why our approach performs better than IR, we qualitatively compare some images from the training set, with respective images generated by IR and by our technique in~\cref{fig:var_imgs}(a-c).
We perform this comparison on CIFAR-100 since its large number of samples per class enables an effective analysis of the variance across sample images of the same class, which we present in the top-row of this figure.
Note that images sampled from IR contain more spatial variance than those from our technique, i.e., our variance images are considerably sharper.
Given that both employ the same augmentations, the reason for this difference is that our generator does not have to learn to reproduce spatial variances that would anyhow be filtered out by $h$, which greatly simplifies the distribution to be learned by the generative model.
Similarly, our replayed samples are expected to contain only the features that are relevant for the classifier, which also enables highly effective knowledge preservation.
Note that this does not constitute a mode collapse in the generative model, i.e., while spatial features exhibit little variance (e.g., rough position and shape of a bottle in CIFAR-100), non-spatial features still can occur with a variance similar to the real images (e.g., color, texture, and background).

\subsubsection{Distillation} \label{sec:abl_dist}
Our framework employs logit and feature distillation for the classifier as well as adversarial distillation for the generative model.
To assess the impact of these contributions, we conduct the following ablation experiments, comparing the metric $\alpha_T$.
Our proposed adversarial distillation is comprised of $\mathcal{L}_G^\mathrm{p}$$=$$\mathcal{L}_{G^\mathrm{ADV}}^\mathrm{p}$$+$$\lambda_\mathrm{ID}\mathcal{L}_{G^\mathrm{ID}}^\mathrm{p}$~(\cref{eq:l_gn_prev}) and $\mathcal{L}_D^\mathrm{p}$~(\cref{eq:l_dn_prev}).
To comparatively study the relative impacts of adversarial and non-adversarial components, we first ablate the image distillation only, called ``Ours\,$-$\,$\mathcal{L}_{G^\mathrm{ID}}^\mathrm{p}$'' in~\cref{tab:aug_dist_ablations}, and then ablate only the adversarial components (i.e., $\mathcal{L}_\mathrm{ADV}^\mathrm{p}$$=$$\mathcal{L}_D^\mathrm{p}$$+$$\mathcal{L}_{G^\mathrm{ADV}}^\mathrm{p}$) called ``Ours\,$-$\,$\mathcal{L}_\mathrm{ADV}^\mathrm{p}$''.
\begin{table}
\caption{Overall accuracies $\alpha_T$ (\%) for ablations of different distillation losses and augmentations, evaluated for 10 tasks on CUB-200.}
    \centering
    \begin{tabular}{llc}
    \toprule
    & Method & $\alpha_T$ \\
    \midrule
    \multirow{3}{*}{Distillation}
    & Ours\,$-$\,$\mathcal{L}_{G^\mathrm{ID}}^\mathrm{p}$         & 41.8 \\
    & Ours\,$-$\,$\mathcal{L}_\mathrm{ADV}^\mathrm{p}$            & 56.2 \\
    & Ours\,$-$\,$\mathcal{L}_\mathrm{FD}$    & 56.8 \\
    \midrule
    \multirow{2}{*}{Augmentation}
    & Ours\,$-$\,CA                           & 56.6 \\
    & Ours\,$-$\,DA                           & 58.3 \\
    \midrule
    & Ours (\gls{ours})                     & \textbf{61.2} \\
    \bottomrule
    \end{tabular}
    \label{tab:aug_dist_ablations}
\end{table}
As seen in~\cref{tab:aug_dist_ablations}, these ablations are subpar compared to our proposed \gls{ours}.
In particular, image distillation has the largest performance impact as omitting $\mathcal{L}_{G^\mathrm{ID}}^\mathrm{p}$ results in a drop of almost 20\,pp in $\alpha_T$.
\begin{figure}[t]
     \centering
     \begin{subfigure}{\linewidth}
         \centering
         \includegraphics[width=\linewidth]{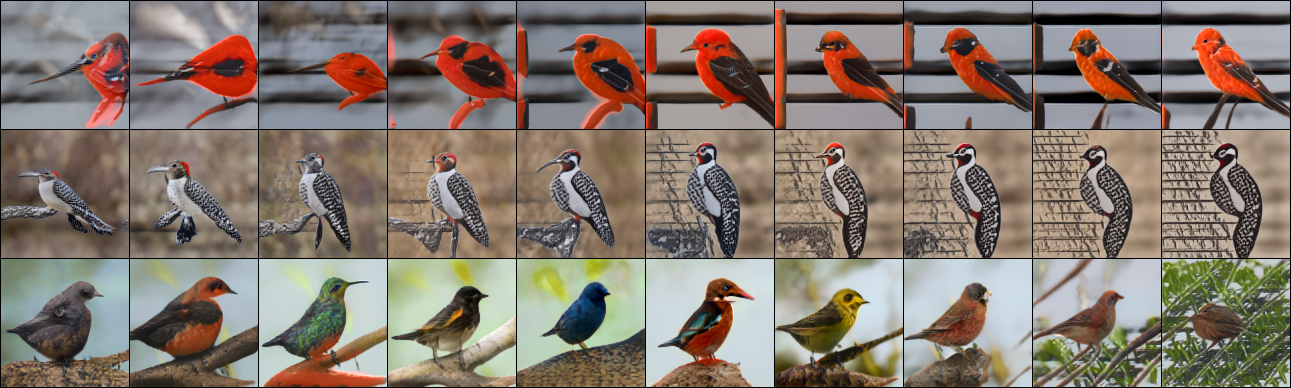}
         \caption{}
         \label{fig:img_evol_noGD}
     \end{subfigure}
     \par\medskip
     \begin{subfigure}{\linewidth}
         \centering
         \includegraphics[width=\linewidth]{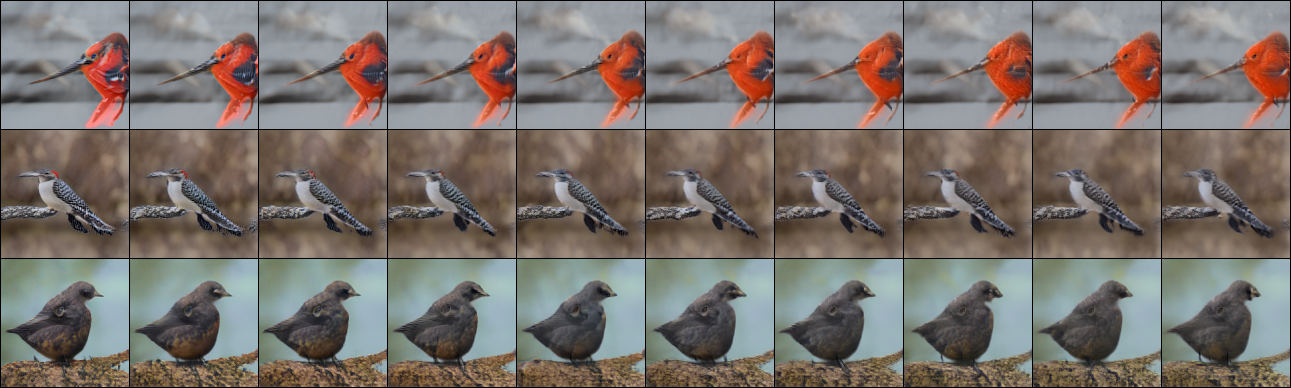}
         \caption{}
         \label{fig:img_evol_noADV}
     \end{subfigure}
     \par\medskip
     \begin{subfigure}{\linewidth}
         \centering
         \includegraphics[width=\linewidth]{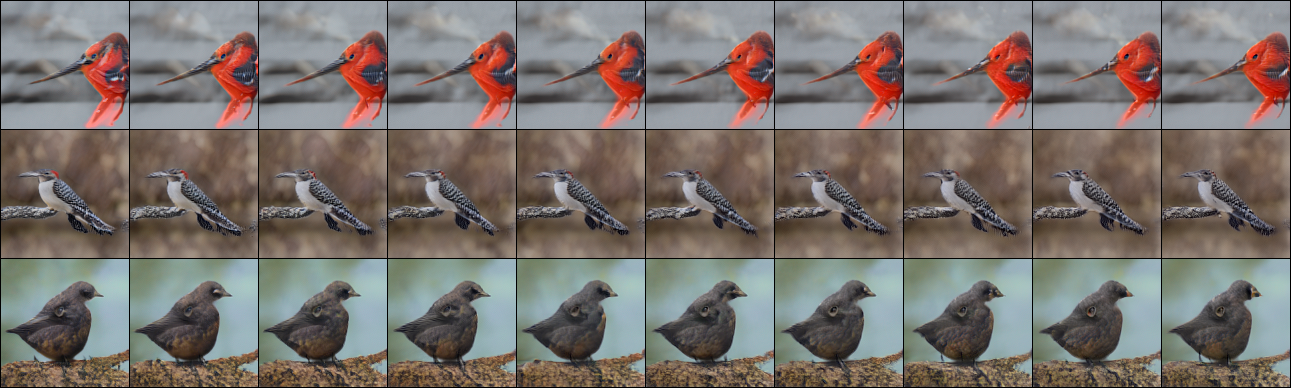}
         \caption{}
         \label{fig:img_evol_ours}
     \end{subfigure}
     \caption{Examples of generated CUB-200 images during a 10-task sequence. Each row shows a different bird class and each column represents a point in the task sequence, starting with $t=1$ in the leftmost column. Images are sampled from (a) a generator without $\mathcal{L}_{G^\mathrm{ID}}$ (b) a generator without $\mathcal{L}_\mathrm{ADV}^\mathrm{p}$ and (c) our proposed generator.}
    \label{fig:img_evol}
\end{figure}

In \cref{fig:img_evol} we provide a qualitative analysis of the impact that the two above distillation components have on generated images.
It is seen in~\cref{fig:img_evol_noGD} that a generator without $\mathcal{L}_{G^\mathrm{ID}}^\mathrm{p}$ suffers from forgetting throughout the task sequence.
With image distillation, forgetting is addressed, but without $\mathcal{L}_\mathrm{ADV}^\mathrm{p}$, the generated images become blurrier as more tasks are learned, as seen upon closer inspection of~\cref{fig:img_evol_noADV}.
In comparison, our proposed generator in~\cref{fig:img_evol_ours} combining both proposed components is able to prevent forgetting while retaining image sharpness, both contributing to the final performance of Genifer.

Finally, an ablation of the feature distillation in the classifier, called ``Ours\,$-$\,$\mathcal{L}_\mathrm{FD}$'' in~\cref{tab:aug_dist_ablations}, shows the substantial benefit also from this component.
Hence, all the proposed distillation objectives are crucial in achieving the state-of-the-art results.

\subsubsection{Effectiveness of Augmentations} \label{sec:abl_augmentations}
In a CIL setting where training data is highly limited, data augmentation is of particular importance. 
To verify this hypothesis, we ablate discriminator augmentation (Ours\,$-$\,DA) and classifier augmentation (Ours\,$-$\,CA) separately in~\cref{tab:aug_dist_ablations}. 
Note that when ablating CA, we still perform conventional augmentations for the current task data, since this is applicable in all replay techniques.
These ablation results show that both augmentations are essential for successful generative replay in CIL.

\section{Conclusion} \label{sec:conclusion}
In this work we have presented a novel framework for class-incremental learning that substantially outperforms the previous state of the art.
Key to our method is a technique to replay images while inheriting the benefits of common feature replay approaches. In contrast to exemplar-based frameworks, our technique does not require storing real samples, which enables application in real-world scenarios with confidentiality or privacy concerns.
Notably, our method also outperforms exemplar-based approaches. 
In ablation studies, we show the value of our proposed feature and adversarial distillation, as well as the importance of augmentations for classifier and GAN training in the context of CIL.
Finally, our novel framework for generative feature-driven image replay furthers the state of the art in CIL, on both CIFAR-100 and CUB-200 in several experimental settings.

\ifCLASSOPTIONcaptionsoff
  \newpage
\fi

\bibliographystyle{IEEEtran}
\bibliography{references}

\begin{thebibliography}{10}
\providecommand{\url}[1]{#1}
\csname url@samestyle\endcsname
\providecommand{\newblock}{\relax}
\providecommand{\bibinfo}[2]{#2}
\providecommand{\BIBentrySTDinterwordspacing}{\spaceskip=0pt\relax}
\providecommand{\BIBentryALTinterwordstretchfactor}{4}
\providecommand{\BIBentryALTinterwordspacing}{\spaceskip=\fontdimen2\font plus
\BIBentryALTinterwordstretchfactor\fontdimen3\font minus
  \fontdimen4\font\relax}
\providecommand{\BIBforeignlanguage}[2]{{%
\expandafter\ifx\csname l@#1\endcsname\relax
\typeout{** WARNING: IEEEtran.bst: No hyphenation pattern has been}%
\typeout{** loaded for the language `#1'. Using the pattern for}%
\typeout{** the default language instead.}%
\else
\language=\csname l@#1\endcsname
\fi
#2}}
\providecommand{\BIBdecl}{\relax}
\BIBdecl

\bibitem{Pan2010ALearning}
S.~J. Pan and Q.~Yang, ``{A survey on transfer learning},'' \emph{IEEE Trans.
  Knowl. Dat. Eng.}, vol.~22, no.~10, pp. 1345--1359, 2010.

\bibitem{McCloskey1989}
M.~McCloskey and N.~J. Cohen, ``{Catastrophic Interference in Connectionist
  Networks: The Sequential Learning Problem},'' in \emph{Psychology of Learning
  and Motivation}, 1989, vol.~24, pp. 109--165.

\bibitem{Goodfellow2013AnNetworks}
I.~J. Goodfellow, M.~Mirza, D.~Xiao, A.~Courville, and Y.~Bengio, ``{An
  Empirical Investigation of Catastrophic Forgetting in Gradient-Based Neural
  Networks},'' \emph{arXiv preprint:1312.6211}, 2013.

\bibitem{Parisi2018ContinualReview}
G.~I. Parisi, R.~Kemker, J.~L. Part, C.~Kanan, and S.~Wermter, ``{Continual
  Lifelong Learning with Neural Networks: A Review},'' \emph{Neural Networks},
  vol. 113, pp. 54--71, 2019.

\bibitem{Farquhar2018TowardsLearning}
S.~Farquhar and Y.~Gal, ``{Towards Robust Evaluations of Continual Learning},''
  \emph{arXiv preprint:1805.09733}, 2018.

\bibitem{vandeVen2019ThreeLearning}
G.~M. van~de Ven and A.~S. Tolias, ``{Three scenarios for continual
  learning},'' \emph{arXiv preprint:1904.07734}, 2019.

\bibitem{DeLange2019ATasks}
M.~De~Lange, R.~Aljundi, M.~Masana, S.~Parisot, X.~Jia, A.~Leonardis,
  G.~Slabaugh, and T.~Tuytelaars, ``A continual learning survey: Defying
  forgetting in classification tasks,'' \emph{IEEE Trans. Pattern Anal. Mach.
  Intell.}, vol.~44, no.~7, pp. 3366--3385, 2022.

\bibitem{masana2020class}
\BIBentryALTinterwordspacing
M.~Masana, X.~Liu, B.~Twardowski, M.~Menta, A.~D. Bagdanov, and J.~van~de
  Weijer, ``Class-incremental learning: survey and performance evaluation on
  image classification,'' \emph{IEEE Trans. Pattern Anal. Mach. Intell.}, pp.
  1--20, 2022. [Online]. Available: \url{https://github.com/mmasana/FACIL}
\BIBentrySTDinterwordspacing

\bibitem{Belouadah2020ATasks}
E.~Belouadah, A.~Popescu, and I.~Kanellos, ``{A Comprehensive Study of Class
  Incremental Learning Algorithms for Visual Tasks},'' \emph{Neural Networks},
  vol. 135, pp. 38--54, 2021.

\bibitem{Rebuffi2017ICaRL:Learning}
S.-A. Rebuffi, A.~Kolesnikov, G.~Sperl, and C.~H. Lampert, ``{iCaRL:
  Incremental Classifier and Representation Learning},'' in \emph{IEEE Conf.
  Comput. Vis. Pattern Recog.}, 2017, pp. 2001--2010.

\bibitem{HouLearningRebalancing}
S.~Hou, X.~Pan, C.~C. Loy, Z.~Wang, and D.~Lin, ``{Learning a Unified
  Classifier Incrementally via Rebalancing},'' in \emph{IEEE Conf. Comput. Vis.
  Pattern Recog.}, 2019, pp. 831--839.

\bibitem{LiuMnemonicsForgetting}
Y.~Liu, Y.~Su, A.-A. Liu, B.~Schiele, and Q.~Sun, ``{Mnemonics Training:
  Multi-Class Incremental Learning without Forgetting},'' in \emph{IEEE Conf.
  Comput. Vis. Pattern Recog.}, 2020, pp. 12\,245--12\,254.

\bibitem{yu2020semantic}
L.~Yu, B.~Twardowski, X.~Liu, L.~Herranz, K.~Wang, Y.~Cheng, S.~Jui, and
  J.~v.~d. Weijer, ``Semantic drift compensation for class-incremental
  learning,'' in \emph{IEEE Conf. Comput. Vis. Pattern Recog.}, 2020, pp.
  6982--6991.

\bibitem{ZhuPASS2021}
F.~Zhu, X.-Y. Zhang, C.~Wang, F.~Yin, and C.-L. Liu, ``Prototype augmentation
  and self-supervision for incremental learning,'' in \emph{IEEE Conf. Comput.
  Vis. Pattern Recog.}, 2021, pp. 5871--5880.

\bibitem{toldoFusion2022}
M.~Toldo and M.~Ozay, ``Bring evanescent representations to life in lifelong
  class incremental learning,'' in \emph{IEEE Conf. Comput. Vis. Pattern
  Recog.}, 2022, pp. 16\,732--16\,741.

\bibitem{Krizhevsky2009LearningImages}
A.~Krizhevsky, ``{Learning Multiple Layers of Features from Tiny Images},''
  Tech. Rep., 2009.

\bibitem{Wah2011TheDataset}
C.~Wah, S.~Branson, P.~Welinder, P.~Perona, and S.~Belongie, ``{The
  Caltech-UCSD Birds-200-2011 Dataset},'' California Institute of Technology,
  Tech. Rep., 2011.

\bibitem{CastroEnd-to-EndLearning}
F.~M. Castro, M.~J. Mar{\'{i}}n-Jim{\'{e}}nez, N.~Guil, C.~Schmid, and
  K.~Alahari, ``{End-to-End Incremental Learning},'' in \emph{Eur. Conf.
  Comput. Vis.}, 2018, pp. 233--248.

\bibitem{Wu2019LargeLearning}
Y.~Wu, Y.~Chen, L.~Wang, Y.~Ye, Z.~Liu, Y.~Guo, and Y.~Fu, ``{Large Scale
  Incremental Learning},'' in \emph{IEEE Conf. Comput. Vis. Pattern Recog.},
  2019, pp. 374--382.

\bibitem{BelouadahScaIL:Learning}
E.~Belouadah and A.~Popescu, ``{ScaIL: Classifier Weights Scaling for Class
  Incremental Learning},'' in \emph{IEEE Wint. Conf. App. Comput. Vis.}, 2020,
  pp. 1266--1275.

\bibitem{Goodfellow2014GenerativeNets}
I.~J. Goodfellow, J.~Pouget-Abadie, M.~Mirza, B.~Xu, D.~Warde-Farley, S.~Ozair,
  A.~Courville, and Y.~Bengio, ``{Generative Adversarial Nets},'' in \emph{Adv.
  Neural Inform. Process. Syst.}, vol.~27, 2014, pp. 2672--2680.

\bibitem{Shin2017ContinualReplay}
H.~Shin, J.~K. Lee, J.~Kim, and J.~Kim, ``{Continual Learning with Deep
  Generative Replay},'' in \emph{Adv. Neural Inform. Process. Syst.}, vol.~30,
  2017.

\bibitem{CongGANForgetting}
Y.~Cong, M.~Zhao, J.~Li, S.~Wang, and L.~Carin, ``{GAN Memory with No
  Forgetting},'' in \emph{Adv. Neural Inform. Process. Syst.}, vol.~33, 2020,
  pp. 16\,481--16\,494.

\bibitem{OstapenkoLearningLearning}
O.~Ostapenko, M.~Puscas, T.~Klein, P.~J{\"{a}}hnichen, and M.~Nabi, ``{Learning
  to Remember: A Synaptic Plasticity Driven Framework for Continual
  Learning},'' in \emph{IEEE Conf. Comput. Vis. Pattern Recog.}, 2019, pp.
  11\,321--11\,329.

\bibitem{WuMemoryForgetting}
C.~Wu, L.~Herranz, X.~Liu, Y.~Wang, J.~Van De~Weijer, and B.~Raducanu,
  ``{Memory Replay GANs: learning to generate images from new categories
  without forgetting},'' in \emph{Adv. Neural Inform. Process. Syst.}, vol.~31,
  2018.

\bibitem{vandeVen2020Brain-inspiredNetworks}
G.~M. van~de Ven, H.~T. Siegelmann, and A.~S. Tolias, ``{Brain-inspired replay
  for continual learning with artificial neural networks},'' \emph{Nat.
  Commun.}, vol.~11, no.~1, pp. 1--14, 2020.

\bibitem{Kemker2017FearNet:Learning}
R.~Kemker and C.~Kanan, ``{FearNet: Brain-Inspired Model for Incremental
  Learning},'' in \emph{Int. Conf. Learn. Represent.}, 2018.

\bibitem{Liu2020GenerativeLearning}
X.~Liu, C.~Wu, M.~Menta, L.~Herranz, B.~Raducanu, A.~D. Bagdanov, S.~Jui, and
  J.~van~de Weijer, ``{Generative Feature Replay For Class-Incremental
  Learning},'' in \emph{IEEE Conf. Comput. Vis. Pattern Recog. Worksh.}, 2020,
  pp. 226--227.

\bibitem{XiangIncrementalNetworks}
Y.~Xiang, Y.~Fu, P.~Ji, and H.~Huang, ``{Incremental Learning Using Conditional
  Adversarial Networks},'' in \emph{Int. Conf. Comput. Vis.}, 2019, pp.
  6619--6628.

\bibitem{Russakovsky2015ImageNetChallenge}
O.~Russakovsky \emph{et~al.}, ``{ImageNet Large Scale Visual Recognition
  Challenge},'' \emph{Int. J. Comput. Vis.}, vol. 115, no.~3, pp. 211--252,
  2015.

\bibitem{perez2017effectiveness}
L.~Perez and J.~Wang, ``The effectiveness of data augmentation in image
  classification using deep learning,'' \emph{arXiv preprint arXiv:1712.04621},
  2017.

\bibitem{Kirkpatrick2017OvercomingNetworks}
J.~Kirkpatrick, R.~Pascanu, N.~Rabinowitz, J.~Veness, G.~Desjardins, A.~A.
  Rusu, M.~Kieran, J.~Quan, T.~Ramalho, A.~Grabska-Barwinska, D.~Hassabis,
  C.~Clopath, D.~Kumaran, and R.~Hadsell, ``{Overcoming catastrophic forgetting
  in neural networks},'' \emph{Proc. Nat. Acad. Sci.}, vol. 114, no.~13, pp.
  3521--3526, 2017.

\bibitem{Nguyen2018VariationalLearning}
C.~V. Nguyen, Y.~Li, T.~D. Bui, and R.~E. Turner, ``{Variational Continual
  Learning},'' in \emph{Int. Conf. Learn. Represent.}, 2018.

\bibitem{Zenke2017ContinualIntelligence}
F.~Zenke, B.~Poole, and S.~Ganguli, ``{Continual Learning Through Synaptic
  Intelligence},'' in \emph{Int. Conf. Mach. Learn.}, 2017, pp. 3987--3995.

\bibitem{MallyaPiggyback:Weights}
A.~Mallya, D.~Davis, and S.~Lazebnik, ``{Piggyback: Adapting a Single Network
  to Multiple Tasks by Learning to Mask Weights},'' in \emph{Eur. Conf. Comput.
  Vis.}, 2018, pp. 67--82.

\bibitem{Serra2018OvercomingTask}
J.~Serr{\`{a}}, D.~Sur{\'{i}}s, M.~Miron, and A.~Karatzoglou, ``{Overcoming
  catastrophic forgetting with hard attention to the task},'' in \emph{Int.
  Conf. Mach. Learn.}, vol.~80, 2018, pp. 4548--4557.

\bibitem{Wortsman2020SupermasksSuperposition}
M.~Wortsman, V.~Ramanujan, R.~Liu, A.~Kembhavi, M.~Rastegari, J.~Yosinski, and
  A.~Farhadi, ``{Supermasks in Superposition},'' in \emph{Adv. Neural Inform.
  Process. Syst.}, vol.~33, 2020, pp. 15\,173--15\,184.

\bibitem{Lesort2020ContinualProcesses}
T.~Lesort, ``{Continual Learning: Tackling Catastrophic Forgetting in Deep
  Neural Networks with Replay Processes},'' \emph{arXiv preprint:2007.00487},
  2020.

\bibitem{Lesort2021RegularizationLearning}
T.~Lesort, A.~Stoian, and D.~Filliat, ``{Regularization Shortcomings for
  Continual Learning},'' \emph{arXiv preprint:1912.03049}, 2019.

\bibitem{lecun1998gradient}
Y.~LeCun, L.~Bottou, Y.~Bengio, and P.~Haffner, ``Gradient-based learning
  applied to document recognition,'' \emph{Proceedings of the IEEE}, vol.~86,
  no.~11, pp. 2278--2324, 1998.

\bibitem{SmithABD2021}
J.~Smith, Y.-C. Hsu, J.~Balloch, Y.~Shen, H.~Jin, and Z.~Kira, ``Always be
  dreaming: A new approach for data-free class-incremental learning,'' in
  \emph{Int. Conf. Comput. Vis.}, 2021, pp. 9374--9384.

\bibitem{Kingma2013Auto-EncodingBayes}
D.~P. Kingma and M.~Welling, ``{Auto-Encoding Variational Bayes},'' \emph{arXiv
  preprint:1312.6114}, 2013.

\bibitem{Mirza2014ConditionalNets}
M.~Mirza and S.~Osindero, ``{Conditional Generative Adversarial Nets},''
  \emph{arXiv preprint:1411.1784}, 2014.

\bibitem{ShenGFROWM2021}
G.~Shen, S.~Zhang, X.~Chen, and Z.-H. Deng, ``Generative feature replay with
  orthogonal weight modification for continual learning,'' in \emph{Int. Joint
  Conf. Neur. Net.}, 2021, pp. 1--8.

\bibitem{CuiDeepCollab2021}
B.~Cui, G.~Hu, and S.~Yu, ``Deepcollaboration: Collaborative generative and
  discriminative models for class incremental learning,'' \emph{AAAI Conf. Art.
  Intell.}, vol.~35, no.~2, pp. 1175--1183, 2021.

\bibitem{Miyato2018CGANsDiscriminator}
T.~Miyato and M.~Koyama, ``{cGANs with Projection Discriminator},'' \emph{arXiv
  preprint:1802.05637}, 2018.

\bibitem{Mescheder2018WhichConverge}
L.~Mescheder, A.~Geiger, and S.~Nowozin, ``{Which Training Methods for GANs do
  actually Converge?}'' in \emph{Int. Conf. Mach. Learn.}, vol.~80, 2018, pp.
  3481--3490.

\bibitem{Hinton2015DistillingNetwork}
G.~Hinton, O.~Vinyals, and J.~Dean, ``{Distilling the Knowledge in a Neural
  Network},'' \emph{arXiv preprint:1503.02531}, 2015.

\bibitem{Karras2020TrainingData}
\BIBentryALTinterwordspacing
T.~Karras, M.~Aittala, J.~Hellsten, S.~Laine, J.~Lehtinen, and T.~Aila,
  ``{Training Generative Adversarial Networks with Limited Data},'' in
  \emph{Adv. Neural Inform. Process. Syst.}, vol.~33, 2020, pp.
  12\,104--12\,114. [Online]. Available:
  \url{https://github.com/NVlabs/stylegan2-ada}
\BIBentrySTDinterwordspacing

\bibitem{Zhao2020ImageTraining}
Z.~Zhao, Z.~Zhang, T.~Chen, S.~Singh, and H.~Zhang, ``{Image Augmentations for
  GAN Training},'' \emph{arXiv preprint:2006.02595}, 2020.

\bibitem{Tran2020OnTraining}
N.-T. Tran, V.-H. Tran, N.-B. Nguyen, T.-K. Nguyen, and N.-M. Cheung, ``{On
  Data Augmentation for GAN Training},'' \emph{IEEE Trans. Image Process.},
  vol.~30, pp. 1882--1897, 2021.

\bibitem{He2016DeepRecognition}
K.~He, X.~Zhang, S.~Ren, and J.~Sun, ``{Deep Residual Learning for Image
  Recognition},'' \emph{IEEE Conf. Comput. Vis. Pattern Recog.}, pp. 770--778,
  2016.

\bibitem{Liu2020OnBeyond}
\BIBentryALTinterwordspacing
L.~Liu, G.~Tech, P.~He, W.~Chen, X.~Liu, J.~Gao, and J.~Han, ``{On the Variance
  of the Adaptive Learning Rate and Beyond},'' in \emph{Int. Conf. Learn.
  Represent.}, 2020. [Online]. Available:
  \url{https://github.com/LiyuanLucasLiu/RAdam}
\BIBentrySTDinterwordspacing

\bibitem{karras2017progressive}
T.~Karras, T.~Aila, S.~Laine, and J.~Lehtinen, ``Progressive growing of gans
  for improved quality, stability, and variation,'' in \emph{Int. Conf. Learn.
  Represent.}, 2018.

\bibitem{Yazc2019TheTraining}
Y.~Yazıcı, C.-S. Foo, S.~Winkler, K.-H. Yap, G.~Piliouras, and
  V.~Chandrasekhar, ``{The Unusual Effectiveness of Averaging in GAN
  Training},'' in \emph{Int. Conf. Learn. Represent.}, New Orleans, LA, USA,
  2019.

\bibitem{Paszke2019PyTorch:Library}
A.~Paszke \emph{et~al.}, ``{PyTorch: An Imperative Style, High-Performance Deep
  Learning Library},'' in \emph{Adv. Neural Inform. Process. Syst.}, vol.~32,
  2019.

\bibitem{verma2021eft}
V.~K. Verma, K.~J. Liang, N.~Mehta, P.~Rai, and L.~Carin, ``Efficient feature
  transformations for discriminative and generative continual learning,'' in
  \emph{IEEE Conf. Comput. Vis. Pattern Recog.}, 2021, pp. 13\,865--13\,875.

\bibitem{Karras2020AnalyzingStyleGAN}
T.~Karras, S.~Laine, M.~Aittala, J.~Hellsten, J.~Lehtinen, and T.~Aila,
  ``{Analyzing and Improving the Image Quality of StyleGAN},'' in \emph{IEEE
  Conf. Comput. Vis. Pattern Recog.}, 2020, pp. 8110--8119.

\end{thebibliography}

\appendices

\section{Training details} \label{app:training_details}

\subsection{CIFAR-100}
This dataset comprises 100 different object categories, split into \numprint{50000} training and \numprint{10000} test samples.
The images are of size 32$\times$32.
We employ a ResNet-18~\cite{He2016DeepRecognition} and  train it on each incremental task for 100 epochs while the learning rate decays from 0.0001 with factor 5 after 30, 60, and 80 epochs.
Due to the small image size, we modify the original ResNet-18 architecture in the first convolutional layer to have kernel size 3$\times$3 instead of 7$\times$7 and a reduced stride of 1.
Furthermore, we do not perform max-pooling after the said layer.
For feature distillation, we employ $\lambda_\mathrm{FD}$$=$$1$.
During classifier training, both the real and the synthetic images are augmented using random horizontal flips.

\subsection{CUB-200}
This dataset consists of \numprint{5994} training and \numprint{5794} test images grouped into 200 classes of different birds.
Using bilinear interpolation, the images are first resized such that the shorter side has length 128, and then upsampled to obtain a shorter side length of 256.
The reason for this procedure is twofold: 
First, we found that a classifier trained on random crops of size 224~$\times$~224 extracted from 256-short-side images yields a competitive upper-bound accuracy on this dataset.
Second, training our generator to synthesize 128~$\times$~128 images is significantly faster than doubling the resolution, without compromising the CL accuracy.
We employ a ResNet-18 (original architecture) and train it on each incremental task for 200 epochs while decaying the learning rate from 0.0001 with factor 5 after 60, 120, and 160 epochs.
For feature distillation, $\lambda_\mathrm{FD}$$=$$0.1$ is used.
Due to the small number of samples per class, we employ both horizontal flips and random crops as augmentations on this dataset.

\subsection{Classifier hyperparameters} \label{app:hyperparam_clsf}

The hyperparameters for the classifier training that are common across both datasets are shown in~\cref{tab:hyperparams_clsf}.
In the setting with a large first task, we tune the logit distillation coefficient for different task splits in the range $\lambda_\mathrm{LD}$$\in$$[0.95, 0.99]$ and report the best results.
In the setting without a large first task, $\lambda_\mathrm{LD}$$=$$0.8$ empirically worked best.
\begin{table}
    \centering
    \caption{Hyperparameters for the classifier training. The batch ratio $\eta$ denotes the ratio between previous and current samples in a batch.}
    \begin{tabular}{lc}
    \toprule
    Batch size & $32$ \\
    Batch ratio $\eta$ & $0.5$ \\ 
    Optimizer & RAdam \\
    Learning rate & $0.0001$ \\
    Weight decay & $0.0005$ \\
    \bottomrule
    \end{tabular}
    \label{tab:hyperparams_clsf}
\end{table}

\subsection{GAN hyperparameters} \label{app:hyperparam_gan}
\begin{table}[t]
\scriptsize
\caption{Number of GAN training iterations for one task in CIFAR-100 (a) and CUB-200 (b).}
\begin{subtable}{\linewidth}
    \centering
    \caption{CIFAR-100}
    \begin{tabular}{lcc}
         \toprule
         & Iterations per task & Classes per task \\
         \midrule
         First task $t=1$ & \num{250000} & $20/40/50$ \\
         \midrule
         \multirow{4}{*}{Tasks $t>1$} & \num{250000} & $20$ \\
         & \num{80000} & $10$ \\
         & \num{40000} & $5$ \\
         & \num{40000} & $3$ \\
         \bottomrule
    \end{tabular}
    \label{tab:hyperparams_gan_cifar}
\end{subtable}
\newline
\vspace*{0.5cm}
\newline
\begin{subtable}{\linewidth}
    \centering
    \caption{CUB-200}
    \begin{tabular}{lcc}
         \toprule
         & Iterations per task & Classes per task \\
         \midrule
         First task $t=1$ & \num{250000} & $100$ \\
         \midrule
         \multirow{3}{*}{Tasks $t>1$} & \num{30000} & $20$ \\
         & \num{30000} & $10$ \\
         & \num{20000} & $5$ \\
        \bottomrule
    \end{tabular}
    \label{tab:hyperparams_gan_cub}
\end{subtable}
\label{tab:hyperparams_gan}
\end{table}
For image distillation, we set $\lambda_\mathrm{ID}$$=$$10 \kappa$, where $\kappa = \frac{|\mathcal{C}_\mathrm{p}|}{|\mathcal{C}_\mathrm{c}|}$ is the ratio between the number of previous and current classes.
For the discriminator augmentation, we empirically found that it is beneficial to limit the probability $p$ of augmentations to $0.5$.
In limited-data scenarios such as ours, $p>0.5$ would otherwise trigger relatively soon and lead to augmentation \emph{leaking}~\cite{Karras2020TrainingData}, e.g., $G$ would start to synthesize upside-down images.

For the discriminator's $R_1$ regularization (see~\cite{Mescheder2018WhichConverge}), we use $\lambda_{R_1}$$=$$0.5$.
Similarly to~\cite{Karras2020AnalyzingStyleGAN}, we employ \emph{lazy} $R_1$ regularization, i.e., $R_1$ is minimized only every $16$ discriminator steps.
We used a batch size of $64$ and an equalized learning rate of $0.0025$ for both discriminator and generator.
Details on the number of iterations in the GAN training are provided in Table~\ref{tab:hyperparams_gan} for each task size.

\section{GAN architecture} \label{app:gan_arch}
\begin{figure*}
     \centering
     \includegraphics[width=\linewidth]{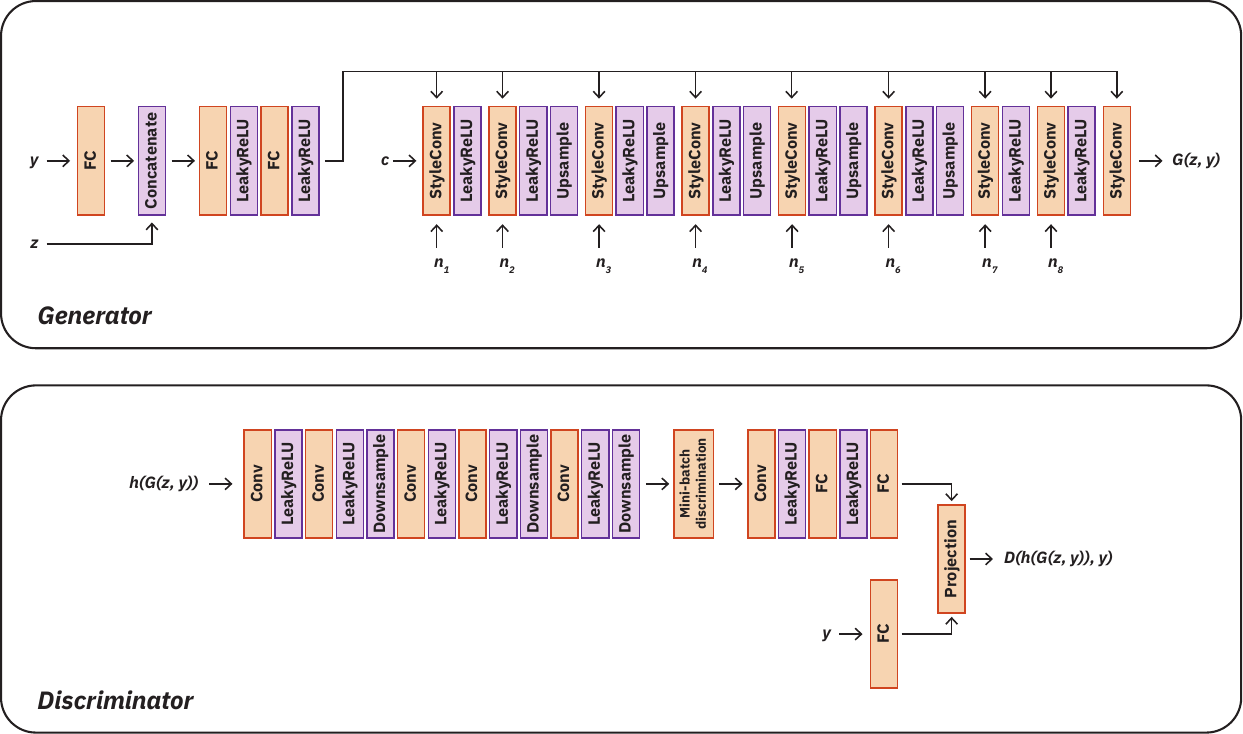}
     \caption{\textbf{Top:} architecture of the generator $G$, where $c$ is a learned input constant, $\mathbf{z}$ is the random noise vector input, and $n_i$ are noise maps that are fed to different layers. $\mathbf{z}$ and $n_i$ are drawn from a standard normal distribution. $y$ is a one-hot vector encoding the class information. A StyleConv module consists of style modulation, convolution, and noise injection. Note that we do not perform any normalization. FC represents fully-connected layers. \textbf{Bottom:} architecture of the discriminator $D$, where the input to $D$ is $h(G(\mathbf{z}, y))$, i.e., the representation (learned by the classifier) of a generated image $G(\mathbf{z}, y)$.}
    \label{fig:gan_arch}
\end{figure*}

The architectures of the generator $G$ and the discriminator $D$ are illustrated in~\cref{fig:gan_arch}.

\section{Qualitative results} \label{app:qual_res}
\begin{figure*}[t]
    \centering
    \includegraphics[width=\linewidth]{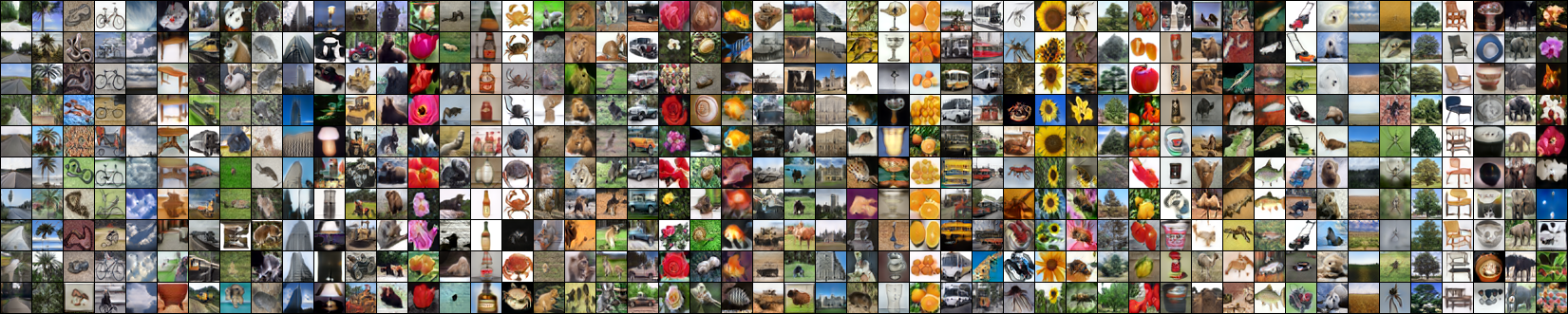}
    \caption{Generated images after training on the first task in CIFAR-100. Each column shows different samples of one class.}
    \label{fig:qual_res_cifar_large}
\end{figure*}

\begin{figure*}[t]
    \centering
    \includegraphics[width=\linewidth]{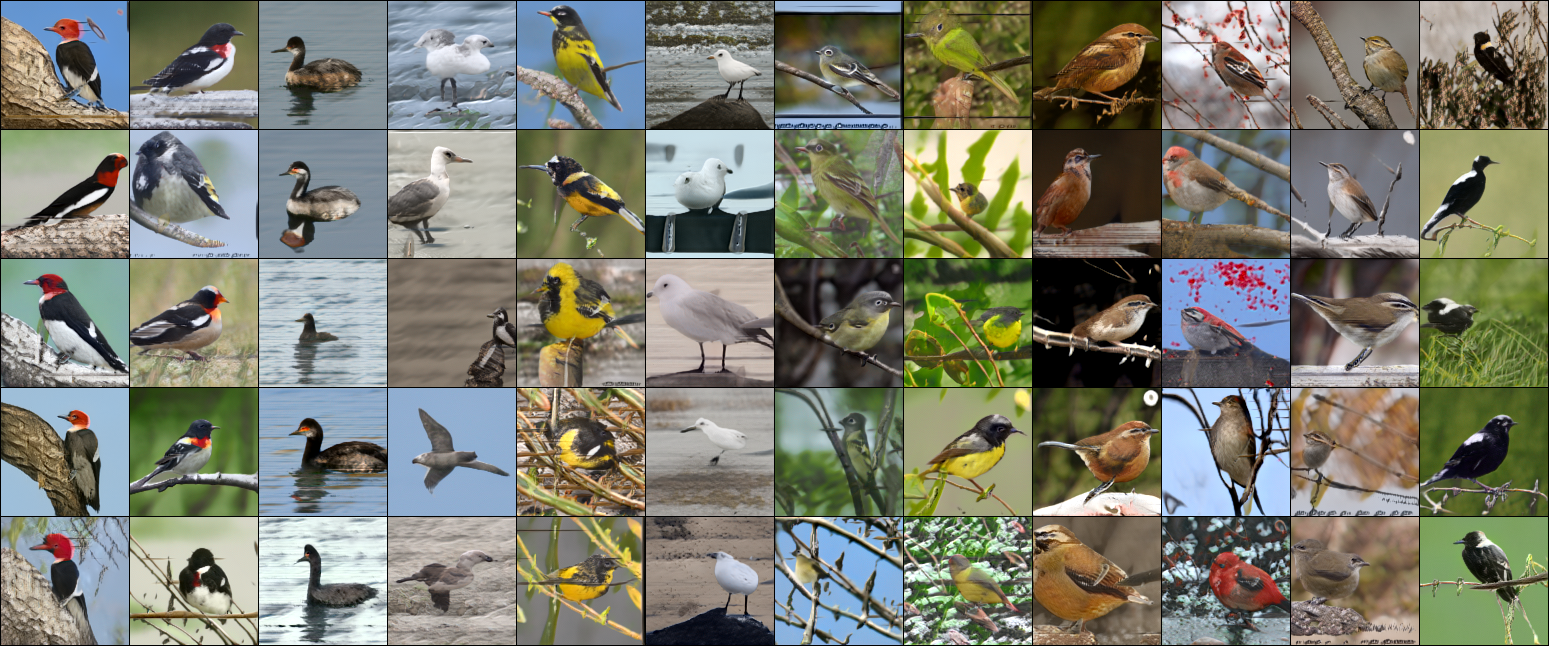}
    \caption{Generated images after training on the first task in CUB-200. Each column shows different samples of one class.}
    \label{fig:qual_res_cub_large}
\end{figure*}

\begin{figure*}[t]
    \centering
    \includegraphics[width=\linewidth]{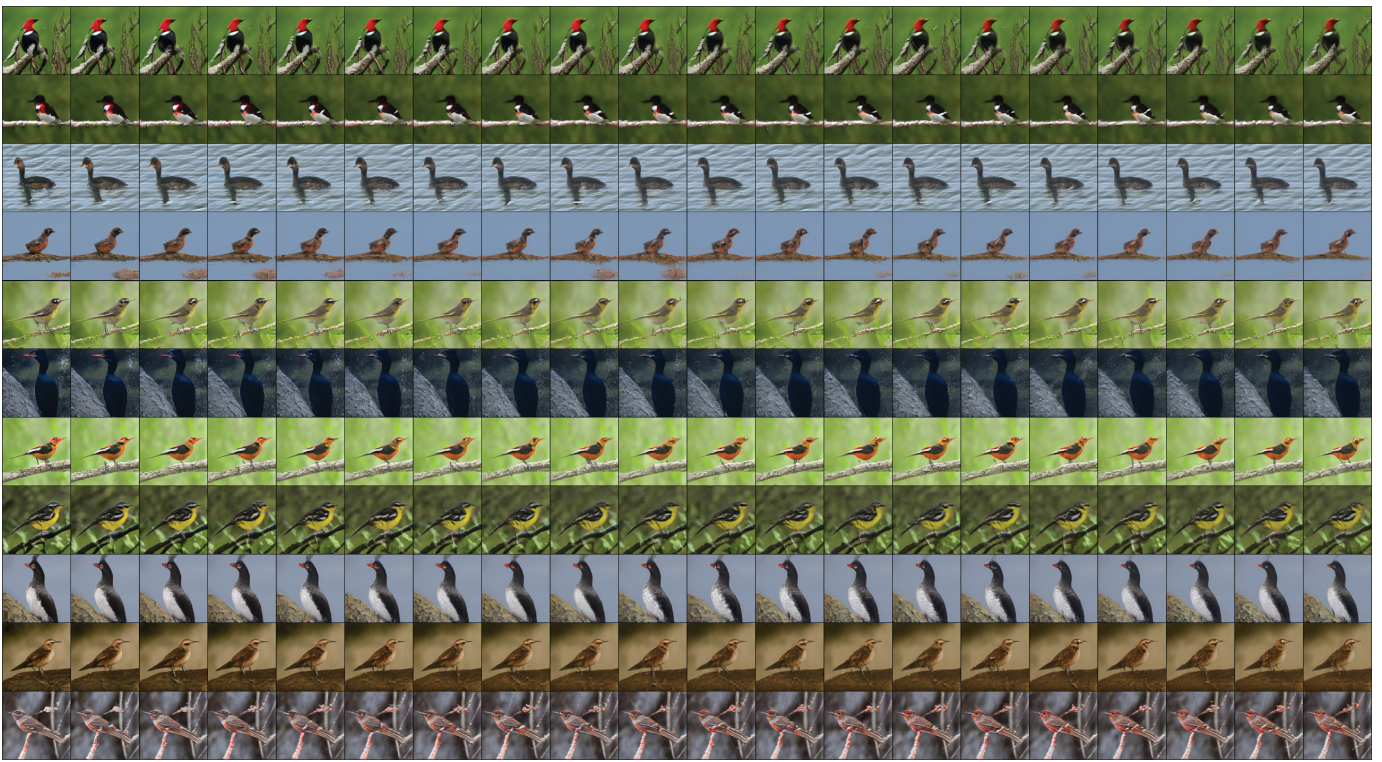}
    \caption{Evolution of generated CUB-200 images. Each row shows how a sample of a class from task $t$$=$$1$ evolves during a sequence. Each column represents the generator's state after learning a task $t$, from $t$$=$$1$ (leftmost) to $t$$=$$19$ (rightmost).}
    \label{fig:qual_res_cub_20x5}
\end{figure*}

\cref{fig:qual_res_cifar_large} and \cref{fig:qual_res_cub_large} show example images replayed by our generator after training on the first large task in CIFAR-100 and CUB-200, respectively.
In~\cref{fig:qual_res_cub_20x5}, we demonstrate how our synthesized samples from the first task evolve while the generator trains on a sequence of 20 tasks in CUB-200.
Only very subtle differences can be observed when comparing the images generated directly after learning the first task and at the end of the sequence.

\end{document}